\pdfoutput=1

\documentclass[11pt]{article}

\usepackage{acl}

\usepackage[T1]{fontenc}
\usepackage{pifont}

\usepackage{times}
\usepackage{soul}
\usepackage{url}
\usepackage[utf8]{inputenc}
\usepackage{graphicx}
\graphicspath{ {.} }
\usepackage{amsmath}
\usepackage{amsthm}
\usepackage{booktabs}
\usepackage{algorithm}
\usepackage{algorithmic}
\usepackage{amsmath}
\usepackage{bbm}
\usepackage{amsfonts}
\usepackage{multirow} 
\usepackage{multicol} 
\usepackage{calc}
\usepackage[disable]{todonotes}
\usepackage{tabularx}
\usepackage{etoolbox}
\usepackage[table]{xcolor}  
\usepackage{pgf}            
\usepackage{xspace}
\usepackage{microtype}
\usepackage{inconsolata}
\usepackage[breakable,skins]{tcolorbox}
\usepackage{latexsym}

\newcommand{\model}[1]{\texttt{#1}\xspace}

\newcommand{\maxSmallTodo}[1]{\todo[size=\tiny]{\textcolor{white}{TODO: }#1}}
\newcommand{\fytodo}[1]{\todo[size=\tiny,color=green]{#1}}

\newcommand{\fydone}[1]{}

\definecolor{LightGray}{rgb}{0.9,0.9,0.9}

\tcbset{
  enhanced,
  boxrule=0.8pt,
  colback=blue!5!white,
  colframe=blue!60!black,
  fonttitle=\bfseries,
}

\newcommand{\src}{\ensuremath{\mathbf{x}}}
\newcommand{\tgt}{\ensuremath{\mathbf{y}}}

\definecolor{frag}{rgb}{0.06,0.45,0.27}
\definecolor{draft}{rgb}{0.10,0.34,0.54}
\definecolor{fragdraft}{rgb}{0.45,0.05,0.47}
\definecolor{instruct}{rgb}{0.7,0.6,0}
\definecolor{baseline}{rgb}{0.6,0.3,0.1}

\newcommand{\Frag}{\textcolor{frag}{(F)}\xspace}
\newcommand{\Draft}{\textcolor{draft}{(D)}\xspace}
\newcommand{\FragDraft}{\textcolor{fragdraft}{(F+D)}\xspace}
\newcommand{\Baseline}{\textcolor{baseline}{(B)}\xspace}
\newcommand{\Instruct}{\textcolor{instruct}{(I)}\xspace}

\newcommand{\BLEU}{\texttt{BLEU}\xspace}
\newcommand{\COMET}{\texttt{COMET}\xspace}
\newcommand{\MetricX}{\texttt{MetricX}\xspace}
\newcommand{\COMETKiwi}{\texttt{COMETKiwi}\xspace}
\newcommand{\sentinelsrc}{\texttt{sentinel-src-25}\xspace}

\newcommand{\green}[1]{\textcolor{green!40!black!60}{#1}}

\title{Machine Translation with Fragment-Based Chain-of-Thoughts}
\title{Fragment-Based Machine Translation}
\title{Reasoning about In-Context Samples for Machine-Translation}

\author{
	Maxime Bouthors$^{\dagger}$ \quad Josep Crego$^{\dagger}$ \quad François Yvon$^{\ddagger}$ \\
	$^{\dagger}$SYSTRAN by ChapsVision, 5 rue Feydeau, F-75002 Paris, France \\
	$^{\ddagger}$Sorbonne Université, CNRS, ISIR, F-75005 Paris, France \\
	\texttt{\{mbouthors,jcrego\}@chapsvision.com} \\
	\texttt{yvon@isir.upmc.fr}
}

\begin{document}

\maketitle

\begin{abstract}
	Large Language Models (LLMs) can be trained to perform chain-of-thoughts reasoning in order to improve the reliability of their responses. 
	In this work, we investigate how explicit reasoning can be leveraged for LLM-Based Machine Translation (MT) with in-context samples. We introduce a novel fragment-based reasoning framework in which the model first extracts parallel source-target fragments from retrieved similar exemplars, and uses these fragments as intermediate reasoning traces to produce the final translation.
	To train our model, we distill silver fragments and drafts from a large teacher model.
	Our experiments with the Qwen3 model family, over 6 languages, including up to 5 domains per language, demonstrate that fragment-based MT significantly outperforms alternative methods like standard k-shot or basic drafting.\footnote{Our code and data are available at \url{https://github.com/Maxwell1447/Fragment-Based-Reasoning}}
\end{abstract}

\begin{figure*}[t]
	\includegraphics[width=\linewidth]{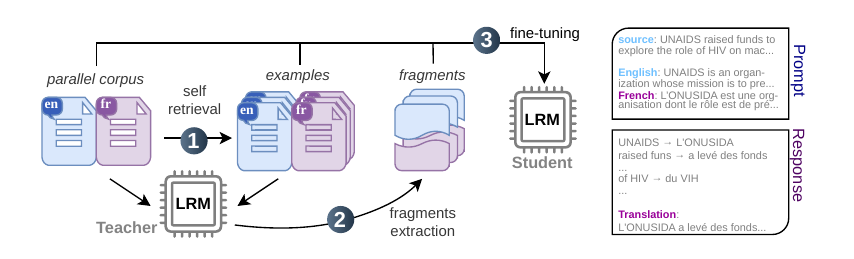}
	\caption{\label{fig:task-pipeline} Overview of the training pipeline of the model: (1) exemplars are retrieved from the TM; (2) A teacher model performs FE from the exemplars; (3) A student model is trained to reproduce FE as a reasoning trace.}
\end{figure*}

\section{Introduction \label{sec:introduction}}

Large Language Models (LLMs) have proven to be successful in a wide variety of tasks \citep{touvron-etal-2023-llama}, incuding Machine Translation (MT) \citep{vilar-etal-2022-prompting,zhang-etal-2022-opt}.
In-context learning (ICL) \citep{brown-etal-2020-language} successfully improves MT-tailored LLMs \citep{zhang-etal-2023-prompting}, as it provides a simple mechanism to input additional task-specific, lexical, terminological or stylistic context to guide the generation algorithm towards more accurate translations \citep{moslem-etal-2023-adaptive}.

LLMs have recently evolved into Large Reasoning Models (LRMs), which first generate ``thinking'' tokens before computing their final output \citep{wei-etal-2022-chain}. LRMs can be further instructed, e.g., through supervised fine-tuning (SFT) or reinforcement learning (RL), to reproduce human reasoning steps in formal domains such as mathematics or coding \citep{deepseekai-2025-deepseekr1}. In the context of MT, these extra tokens serve, for instance, to simulate linguistic analyses, partial translations or drafts before generating the final translation \citep{raunak-etal-2023-leveraging,he-etal-2024-exploring,briakou-etal-2024-translating,zebaze-etal-2025-reasoning}.

Most works on Chain-of-Thoughts (CoT) for MT thus take inspiration from the activities of human translators, modeled as a series of logically organized steps, from dictionary search to terminological analysis, translation and then revision(s), and make their models simulate the corresponding ``thinking'' tokens. Our inspiration is different, as we would like to model the reasoning process associated with the edition of one or several close exemplar(s) in translation-memory (TM) augmented translation \citep{bowker-fisher-2010}. Despite the advances of NMT systems, TMs remain important tools in professional translators' workbench, as they enable the \emph{transparent} reuse of high-quality translations, which have already been validated for their correct terminology and phraseology -- two key aspects in specialized translation. The integration of TMs with NMT, then with in-context learning and LLMs, is thus an important issue, studied e.g., in \citep{gu-etal-2018-search,bulte-tezcan-2019-neural}.

As discussed in early works on Example-Based Machine Translation (EBMT) \cite{somers-1999-review}, reasoning with TMs can also logically be decomposed in several steps: (a) retrieving relevant \textsl{exemplars}\footnote{Exemplars are instances of past translations, stored in a TM and associating a source text and a target side translation.} for the current source sentence, typically based on fuzzy surface similarity scores; (b) matching segments in the retrieved exemplars that also occur in the source sentence and retrieving their target side equivalent; (c) selecting, recombining, adapting and completing these segments into a translation; (d - optional) revising this draft translation.

In this work, our main focus is on steps (b), and to a lesser extent (c). We rely on external retrieval modules for (a), and on the built-in generation abilities of LLMs for (d). We accordingly ask the following research questions:
(RQ1) can LLMs/LRMs reliably perform the matching step and identify, rather than generate, actual translation fragments in retrieved exemplar(s)? (RQ2) Are the corresponding fragments improving the MT quality? (RQ3) Does fragment extraction (henceforth FE) benefit more when multiple exemplars are retrieved? In other words, can they sort valuable information from noise? (RQ4) Does FE benefit more for some specific domains or languages than for others?\todo{Introduce the concept of Example-Based Reasoning for MT - EBRMT}

To study these questions, we develop a complete pipeline, illustrated in Figure~\ref{fig:task-pipeline},\todo{use exemplars instead of examples; improve caption; where do we introduce Qwen?} relying on the Qwen3-32B model \citep{yang-etal-2025-qwen3} as a teacher to generate training samples for the matching task, and on a finetuned version of Qwen3-8B to perform the reasoning and translation generation tasks. Our main findings are the following:
(a) LLMs can learn to match, select or generate useful parallel fragments in parallel sentences;
(b) Extracting such fragments during the reasoning process -- even when noisy -- significantly improves the translation quality for all metrics considered in this work;
(c) The gains are independant of the number of retrieved exemplars, suggesting that FE produces a robust signal that can be exploited to generate the final translation;
(d) This method generalizes to domains that are unseen during fine-tuning;
(e) Combining segment retrieval with drafting techniques seems to undermine translation quality for domains seen in training, but yields improvement for unseen domains.



\section{Related Work \label{sec:related}}

\paragraph{Retrieval Augmented Machine Translation}

Retrieval-augmented MT (RAMT) leverages similar exemplars to improve translation quality. As for other cognitive tasks, relying on exemplars which can be inspected increases the transparency of the translation process \citep{rudin-cynthia-2019-stop}; for specialized MT, exemplars also implicitly provide some domain adaptation capabilities. RAMT is readily implemented in encoder-decoder models by augmenting the source-side with one or multiple similar target sentence(s) retrieved from the TM, while the target side decoder module remains mostly unchanged, as proposed in \citep{bulte-tezcan-2019-neural,xia-etal-2019-graph,he-etal-2021-fast,cheng-etal-2022-neural,agrawal-etal-2023-context},

LLM-based MT also bode well with exemplars: once integrated through in-context learning (ICL), they provide task-specific contexts along with lexical and stylistic suggestions \citep{radford-etal-2019-language}.
Multiple follow-up works have further explored the impact of the prompt, of the quality and number of exemplars and of the retrieval procedure \citep{moslem-etal-2023-adaptive,vilar-etal-2023-prompting,zhang-etal-2023-prompting,hendy-etal-2023-howgood,bawden-yvon-2023-investigating,zebaze-etal-2025-context}.

A parallel strand of research focuses on \emph{edit-based models} \citep{gu-etal-2019-levenshtein}, viewing RAMT as a new instance of example-based NMT \citep{nagao-1984-framework,somers-1999-review,carl-etal-2004-recent}. Contrary to NMT or LLM-based approaches, where the output is computed from scratch, the output is computed by minimally patching an existing translation. This requires to identify the parts that respectively need to be kept, adapted, or retranslated from scratch, then to perform the prescribed operations. Such methods are especially effective when very close exemplars, necessitating a small number of edits, can be found; in this way, the output is very likely to contain large, error-free, fragments of the retrieved translation. Recent implementations of these ideas, relying on non-autoregressive decoders, are presented in e.g., \citep{xu_editor_2021,niwa-etal-2022-nearest,xu-etal-2023-integrating,zheng-etal-2023-towards,bouthors-etal-2023-towards}.

\paragraph{Large Reasoning Models}

Large Reasoning Models (LRMs) are able to perform reasoning steps and solve complex tasks \citep{cobbe-etal-2021-training}. These models are tuned to perform additional steps of intermediary text generation, dubbed ``chain-of-thoughts'' (CoT) steps, before generating their final answer \citep{wei-etal-2022-chain}. CoT can be toggled via prompting ("\textit{Let us think step by step}") or supervised fine-tuning \citep{kojima-etal-2022-large,zhang-etal-2023-automatic,yasunaga-etal-2024-large}. RL can further improve such reasoning capabilities, especially for code or problem-solving tasks where the validity of answers can be automatically checked \citep{setlur-etal-2024-rewarding,deepseekai-etal-2025-deepseek}.
%

\paragraph{Large Reasoning Machine Translation Models}

\citet{raunak-etal-2023-leveraging} explore various ways to perform automatic post-editing through prompting GPT-4, contrasting prompts for (a) generation of a refined translation, (b) generation of a list of edits, then of the revision (dubbed \emph{CoT}), (c) generation of MQM-like\footnote{MQM is the Multidimensional Quality Metrics framework of \citet{burchardt-2013-multidimensional}.} annotations, then edits, then of the revision (\emph{structured CoT}). In their setting, CoT-based refinements underperform the baseline, yet produce valuable lists of possible edits. Work on iterative refinement is continued, e.g., by \citet{feng-etal-2024-tear, xu-etal-2024-llmrefine,chen-etal-2024-iterative}.

In contrast to these works, which focus on post-generation steps, MAPS (Multi-Aspect Prompting and Selection, \citep{he-etal-2024-exploring}) focuses on \emph{pre-translation}. This approach prompts an LLM to analyze the source sentence and generate a list of topics, of keywords, and demonstrations in the form of similar exemplars. This background information is then used to produce three candidate translations, each conditioned on a different knowledge source. The most promising candidate, as evaluated by a quality estimation (QE) metric, is finally selected. 

\citet{briakou-etal-2024-translating} combine both viewpoints, integrating subtasks that translators perform prior and posterior to translation. Their experiments use one pre-translation (explain ``difficult'' source terms), and three post-translation (draft, refine, proofread) steps, focussing on document'-level MT.\footnote{In practice, segments of maximum 250 tokens obtained by merging adjacent sentences.} 
This approach is extended by \citet{he-etal-2025-r1t1}, who simulate various ``reasoning'' strategies used by translators (e.g., decomposition, back translation, pivot translation), combining SFT and RL.

\citet{chen-etal-2025-evaluating,nguyen-xu-2025-reasoning} more directly attempt to evaluate ``reasoning'' models for MT using prompting techniques.
\citet{feng-etal-2025-mtr1} reuse training strategies of DeepSeek-R1 \citep{deepseekai-etal-2025-deepseek} to tune MT-adapted LRMs. For this, they study several reward models, combining format and content conformity metrics. Their results suggest that training an effective MT engine with pure RL pays off. 
\citet{zebaze-etal-2025-reasoning} implement CoT via pure SFT and find that standard CoT -- i.e., allowing ``thinking'' tokens -- is not helping MT performance in their setting.
Two types of CoT are considered: (a) translation introspection, simulating the reasoning process of a trained translator in 6 different ways \citep{he-etal-2025-r1t1}; (b) auxiliary tasks, where CoT tokens correspond to tasks that can help translation, termed ``modular prompting strategies''. These various ``pre-translation'' tasks, such as paraphrasing the source, or identifying translation difficulties, are distilled from teacher models, then assembled into SFT examples for the student model. Similar experiments are reported by \citet{rajaee-etal-2026-unlocking}.



\begin{figure}[t]
	\includegraphics[width=\linewidth]{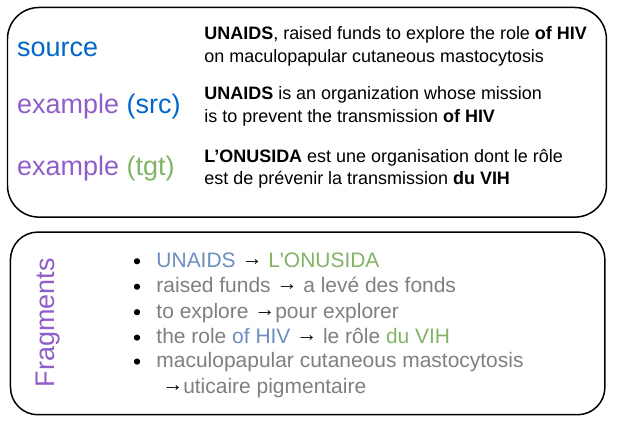}
	\caption{\label{fig:task-illustration} Illustration of the fragment extraction (FE) step, with $k=1$ exemplar. Extracted fragments are colored while generated parts are grayed.}
\end{figure}

\section{Method \label{sec:method}}

\subsection{Principles \label{ssec:principles}}

From a bird's-eye view, EBMT is characterized by three main operations: retrieval, matching and recombination \citep{somers-1999-review}. Optionally, a revision step can finally be applied. In this study, we mostly focus on the second and third steps. Given relevant exemplars retrieved from memory, the matching step first locates source chunks that are relevant for the current translation task, then matches these with the corresponding target chunks. The recombination step then reassembles these target fragments to form a complete translation hypothesis. During this step, some adaptation (e.g., morphological changes, local reorderings, etc) may be required to ensure the syntactic well-formedness of the hypothesis.

We approach the training of a translation system capable of simulating these processes as follows: we first prompt a strong \emph{teacher model} to generate reasoning traces from inputs made of the source sentence $\src$, augmented with $k$ relevant exemplars $\{(\src_1, \tgt_1), \dots, (\src_k, \tgt_k)\}$ retrieved from a TM. Details are in \textsection~\ref{ssec:silver-fragm-extr}. Once (artificial) reasoning traces are available for each training sample, we fine-tune a \emph{student model} in a supervised manner to reproduce similar reasoning steps during the translation process (see \textsection~\ref{ssec:fragm-based-transl}).
\maxSmallTodo{Expliquer pk on a pas la référence ?}

This training pipeline heavily relies on automatically generated ``reasoning'' tokens,  as is custom in CoT approaches \citep{shum-etal-2023-automatic,zhang-etal-2023-automatic}. This is because these tokens are rarely observed and available in the training data, except in very specific cases (e.g., mathematical problems). Machine Translation is no exception to this state-of-affair, prompting us to simulate translators' reasoning operations. In the next sections, we first document the teacher, then the student models.

\subsection{Silver fragments extraction \label{ssec:silver-fragm-extr}}

Given a source sentence $\src$ and a set of $k$ similar exemplars $E(k) = \{(\src_1, \tgt_1), \dots, (\src_k, \tgt_k)\}$ retrieved from a TM, we would like to generate traces of the following sequential reasoning process:
\begin{enumerate}
\item Decompose $\src$ into a set of minimal translatable semantic units $(u_1, \dots, u_m)$ forming an ordered partition of $\src$, with potentially missing punctuations or function words;
\item For each unit $u_i$, propose a translation $v_i$ that is either (a) a translation of $u_i$ attested in at least one exemplars, or (b) an adaptated version of a segment resembling $u_i$ or (c) novel translation generated from scratch;
\item Compute a draft translation $\tilde \tgt$ obtained by recombining the $v_i$'s together, with possible additional inserts or reformulations to fill the gaps between the $v_i$'s.
\end{enumerate}
Steps~1 and 2 are illustrated in Figure~\ref{fig:task-illustration}. \fydone{'on' is misssing in the figure --> c'est normal que des ponctuations et des mots de liasons puissent sauter}
Each of these steps is reminiscent of processes already well studied in MT systems from previous generations. For instance, step~1 and 2 are similar to the generation of a (weightless) translation table in statistical MT systems, while step 3 is akin to decoding with this table \cite{koehn-2010-smt}. Step~1 and 2 are also standard in EBMT, where step~1 would resort to some sort of syntactic parsing, while step~2 would use lexical access or automatic word alignment. Drafting (step~3) is used in several contemporary MT reasoning models e.g., \cite{xu-etal-2024-llmrefine,briakou-etal-2024-translating}. Focusing on low-resource MT, \cite{zebaze-etal-2025-compositional} also combines steps~1-3, generating translations of short fragments, that are then used as in-context samples for the full sentence. 

The difficulties of these steps are thus well known: Step~1 is ill-defined and ambiguous, as there may be multiple ways to decompose $\src$. Step~2 is also very difficult, as it implies the computation of fuzzy alignments between source and target units, then the selection of one target translation for each source unit, whenever several are available (see e.g., \citet{bouthors-etal-2023-towards} for an attempt to implement these computations with conventional alignment techniques). Finally, step~3 might also imply significant structural changes to reassemble the extracted fragments.


Rather than implementing each step with a dedicated tool, we resort here to prompting and ask a strong teacher LLM (as indicated in Appendix~\ref{appendix:qwen-fragments-prompting}) to successively generate the relevant set of fragments $F = \{ (u_1, v_1), \dots, (u_n, v_n) \}$, which we call \emph{silver fragments}, and a \emph{draft} translation $\tilde \tgt$. After applying this process to the entire parallel training corpus, we obtain annotations that can be used to extend the training data with additional supervision traces of a simulated ``reasoning'' process.

\subsection{Translating with Example-Based Reasoning\label{ssec:fragm-based-transl}}

Using the artificial traces computed by the teacher model, we fine-tune a student LLM to augment the generation process with similar ``thinking'' tokens before computing its final translation $\tgt$.
Denoting $F$ the set of fragments $\{ (u_1, v_1), \dots, (u_m, v_m) \}$, we adapt the parameters $\theta$ of an existing model to compute:\footnote{Notwithstanding additional context information, e.g., system instructions, embedded in the prompt.}
\begin{equation*}
  p_{\theta}(F \circ \tilde \tgt \circ \tgt | \src ; ((\src_1, \tgt_1), \dots, (\src_k, \tgt_k))),
\end{equation*}
where $\circ$ is the concatenation operator. This model is refered to as \textbf{Fragments+Draft}, abbreviated as \FragDraft below.

For comparison purposes, we also fine-tune the following variants on the same dataset, where we selectively ablate sections of the reasoning traces:
\begin{itemize}
\item {Fragments only} \Frag:\\ $p_{\theta}(F \circ \tgt | \src ; ((\src_1, \tgt_1), \dots, (\src_k, \tgt_k)))$,
\item {Draft only} \Draft:\\ $p_{\theta}(\tilde y \circ \tgt | \src ; ((\src_1, \tgt_1), \dots, (\src_k, \tgt_k)))$,
\item {Baseline} \Baseline:\\ $p_{\theta}(\tgt | \src ; ((\src_1, \tgt_1), \dots, (\src_k, \tgt_k)))$. 
\end{itemize}

For all these models, adaptation is performed by optimizing the cross-entropy loss on the artificial fine-tuning data. Details regarding the fine-tuning process and the prompts are in Appendix~\ref{appendix:qwen-translation-prompting}.
\fydone{Je ne comprends pas les prompts dans les figures 5-8. En particulier la partie user est toujours la même alors que j'aurais pensé qu'elle contiendrait explicitement l'instruction de générer les fragments ou le draft selon les cas.}

\section{Data and Metrics \label{sec:datametrics}}

\subsection{Data \label{ssec:datasets}}

The corpus used for the experiments comprises five language pairs, associating English with a European language. All these data are publicly available on the OPUS website\footnote{\url{https://opus-nlpl.eu}} \citep{tiedemann-2012-parallel}. It includes the five English-German domains used by \citet{aharoni-goldberg-2020-unsupervised} and five of the English-French domains of \citet{xu-etal-2022-boosting}. Additionally, we use three English-Polish, two English-Ukrainian and one English-Spanish domains.
These corpora are filtered, and split into train/dev/test when necessary.

For each of the 16 domains and language pairs, a subset of high-quality 10k samples is selected for training and concatenated into a multilingual dataset of size 160k. For each training instance, we uniformly sample between $k=0$ and $k=3$ examples by retrieving in-domain similar sentences from the pool of all parallel sentences.\footnote{We exclude dev and test splits, and retrieve from the whole training set, not just the 10k subsets.} The retrieval setting leverages both BM25 and the Levenshtein distance, as advised by \cite{bouthors-etal-2024-retrieving}.
We reserve a distinct test set (respectively development set) of 1,000 (resp.\ 100) examples for each domain, composing a multi-domain testset of 16k samples.
Additionally, we consider a ``surprise'' English-French test set GNOME, which is not included in the training data, to evaluate the generalization capabilities of the method.

Details regarding the data are in Appendix~\ref{appendix:data}.


\subsection{Metrics\label{ssec:metrics}}

We assess machine translation quality with \BLEU{} \citep{papineni-etal-2002-bleu} computed with SacreBLEU \citep{post-2018-call},\footnote{signature: \texttt{nrefs:1|case:mixed|eff:no|tok:13a| smooth:exp|version:2.1.0};} as well as \COMET\footnote{\url{Unbabel/wmt22-comet-da}, using the defaults settings.} \citep{rei-etal-2020-comet}, and MetricX\footnote{\url{google/metricx-24-hybrid-large-v2p6-bfloat16}} \citep{juraska-etal-2024-metricx}.



\section{Experimental Settings \label{sec:experiments}}

\subsection{Models \label{ssec:models}}

The teacher model mentioned in Section~\ref{ssec:silver-fragm-extr} is \model{Qwen3-32B},\footnote{\url{https://huggingface.co/Qwen/Qwen3-32B}.} which we query with the prompt described in Appendix~\ref{appendix:qwen-fragments-prompting}. As for the student model, we use a smaller model, \model{Qwen3-8B}.\footnote{\url{https://huggingface.co/Qwen/Qwen3-8B}.} Note that the prompt used for the teacher model is quite long, as it includes detailed instructions and an illustration of the task. The use of a smaller student with a minimal prompt is motivated by the computational cost and constraints of deploying or fine-tuning a large model at scale, as well as by the desire to demonstrate that the proposed method can be effective even with a small model and a simple prompt.

Each model equally observes samples with $k=0, 1, 2, 3$ exemplars during training. Notably, 20\% of the training samples have an empty reasoning trace, thereby allowing us to perform inference in two different ways: with thinking enabled vs.\ thinking disabled. These two inference modes are compared below. All fine-tuning processes run for 2~epochs. Detailed settings are in Appendix~\ref{appendix:qwen-ft}.

\subsection{Baselines \label{ssec:baselines}}

In addition to the fine-tuned baseline model \Baseline, with which we generate the translation without any thinking process,\footnote{This model is trained to only generate a translation given the source and the potential exemplars.} we also consider the instruct Qwen3-8B model without fine-tuning, which we denote \Instruct. Since its reasoning\footnote{The reasoning mode of \Instruct (prior to SFT) is very different from that of its fine-tuned counterparts. In particular, the associated traces are much longer, suggesting that the ``thinking'' modes of these models cannot really be compared.} often requires thousands of tokens, it is very costly to run on our large multi-domain test set. Therefore, we only use it with the reasoning ability turned off. 

\begin{table*}
	\centering
	\begin{tabular}{llrrrrrrrrrrrr}
		\toprule
		      &                   & \multicolumn{4}{c}{\BLEU$\uparrow$} & \multicolumn{4}{c}{\COMET$\uparrow$} & \multicolumn{4}{c}{MetricX$\downarrow$}                                                                                                             \\
		\cmidrule(lr){3-6} \cmidrule(lr){7-10} \cmidrule(lr){11-14}
		Model & think             & k=0                                 & k=1                                  & k=2                                     & k=3       & k=0       & k=1       & k=2       & k=3       & k=0       & k=1       & k=2       & k=3       \\
		\cmidrule(lr){1-2}\cmidrule(lr){3-6} \cmidrule(lr){7-10} \cmidrule(lr){11-14}
		\Instruct     & \ding{55}         & 28.6                                & 33.0                                 & 35.6                                    & 36.0      & 82.3      & 79.9      & 81.9      & 82.1      & 7.43      & 2.15      & 2.12      & 2.10      \\
		\Baseline     & \ding{55}         & 38.3                                & 45.3                                 & 46.0                                    & 46.3      & 85.3      & 86.3      & 86.4      & 86.5      & 2.27      & 2.11      & 2.09      & 2.07      \\
		\cmidrule(lr){1-2}
		\Draft     & \ding{55}         & 36.7                                & 44.7                                 & 45.5                                    & 45.8      & 85.0      & 86.1      & 86.2      & 86.4      & 2.36      & 2.16      & 2.13      & 2.11      \\
		\Draft     & \green{\ding{51}} & 36.8                                & 44.1                                 & 45.0                                    & 46.0      & 84.9      & 85.0      & 85.6      & 85.6      & 2.34      & 2.66      & 2.43      & 2.11      \\
		\cmidrule(lr){1-2}
		\Frag     & \ding{55}         & 37.4                                & 44.8                                 & 45.5                                    & 45.9      & 85.1      & 86.2      & 86.3      & 86.4      & 2.33      & 2.13      & 2.11      & 2.09      \\
		\Frag     & \green{\ding{51}} & \bf{38.5}                           & \bf{45.7}                            & \bf{46.4}                               & \bf{46.7} & \bf{85.6} & 86.3      & \bf{86.7} & \bf{86.7} & \bf{2.16} & \bf{2.03} & \bf{1.98} & \bf{1.99} \\
		\cmidrule(lr){1-2}
		\FragDraft   & \ding{55}         & 37.5                                & 44.7                                 & 45.5                                    & 45.8      & 85.0      & 86.1      & 86.3      & 86.4      & 2.34      & 2.15      & 2.11      & 2.10      \\
		\FragDraft   & \green{\ding{51}} & 38.2                                & \bf{45.6}                            & \bf{46.3}                               & \bf{46.6} & \bf{85.5} & \bf{86.5} & \bf{86.6} & 86.6      & \bf{2.20} & \bf{2.04} & \bf{2.01} & \bf{2.01} \\
		\bottomrule
	\end{tabular}
	\caption{\label{tab:main-results} Average \BLEU, \COMET, and \MetricX scores on the multi-domain test set for the different models and number of examples $k$. Note that fine-tuned models can be run with or without reasoning. Results significantly better (p<1\%) than the baseline \Baseline are in \textbf{bold}. The significance is assessed with a paired t-test for \COMET and \MetricX, and with bootstrapping for \BLEU with SacreBLEU (n=1000).}
\end{table*}

\section{Results \label{ssec:results} and Analysis}

\subsection{Fragment-based MT}

We first study the translation quality (RQ1) of various configurations (fragments, drafting) and its dependency to the number of retrieved exemplars $k$. The main results are reported in Table~\ref{tab:main-results}. A clear trend across all metrics indicates that the use of fragments significantly improves the translation quality, with or without drafting (\Frag and \FragDraft). Detailed results broken down by domains / languages are in Appendix~\ref{appendix:perdomain-scores}, see also \textsection\ref{ssec:perdomains}.

A second observation is that drafting alone seems to degrade the translation quality (both \Draft vs. \Baseline and \FragDraft vs. \Frag). This is somewhat unexpected, as drafting is supposed to provide a useful intermediate step for the final translation. This may be because the draft is not sufficiently accurate, and thus provides a noisy signal for the final translation, or because the model is not able to properly leverage this information. However, in most cases, the draft is a mere combination of the previously generated silver fragments, which clearly contribute to the improvement. See discussion in \textsection\ref{ssec:silver-infered}.
Finally, we find that the gains provided by reasoning are observed for all values of $k$, which suggests that the model can successfully benefit from fragments of sufficient quality.

Table~\ref{tab:winrate-comet} displays an alternative view of these results via the win rate of \Frag against \Baseline with respect to three metrics across our 16~domains. Fragment-based translation outperforms the baseline in 14 out of 16 cases on average, with no clear dependency on the number of exemplars $k>0$. 

\begin{table}[ht]
	\begin{tabular}{lrrrr}
		\toprule
		$k=$     & 0     & 1     & 2     & 3     \\
		\cmidrule(lr){2-5}
		\BLEU    & 8/16  & 12/16 & 14/16 & 14/16 \\
		\COMET   & 14/16 & 14/16 & 14/16 & 14/16 \\
		\MetricX & 6/16  & 13/16 & 15/16 & 14/16 \\
		\bottomrule
	\end{tabular}
	\caption{
		\label{tab:winrate-comet} Win rates of \Frag (with thinking on) against \Baseline using \BLEU, \COMET, and \MetricX across 16~domains.}
\end{table}

\begin{table*}
	\centering
	\begin{tabular}{llrrrrrrrrrrrr}
		\toprule
		      &                   & \multicolumn{4}{c}{\BLEU$\uparrow$} & \multicolumn{4}{c}{\COMET$\uparrow$} & \multicolumn{4}{c}{MetricX$\downarrow$}                                                                                                \\
		\cmidrule(lr){3-6} \cmidrule(lr){7-10} \cmidrule(lr){11-14}
		Model & think             & k=0                                 & k=1                                  & k=2                                     & k=3      & k=0      & k=1      & k=2      & k=3      & k=0      & k=1      & k=2      & k=3  \\
		\cmidrule(lr){1-2}\cmidrule(lr){3-6} \cmidrule(lr){7-10} \cmidrule(lr){11-14}
		\Instruct     & \ding{55}         & 44.6                                & 55.7                                 & 56.9                                    & 57.4     & 85.6     & 84.3     & 85.8     & 85.7     & 1.68     & 2.21     & 2.01     & 2.07 \\
		\Baseline     & \ding{55}         & 45.9                                & 64.4                                 & 65.2                                    & 65.4     & 86.0     & 89.4     & 89.8     & 89.8     & 1.65     & 1.27     & 1.22     & 1.20 \\ \cmidrule(lr){1-2}
		\Draft     & \ding{55}         & 44.9                                & 64.5                                 & \bf 65.5                                & \bf 65.8 & 85.8     & 89.2     & 89.8     & 89.9     & 1.74     & 1.31     & 1.26     & 1.26 \\
		\Draft     & \green{\ding{51}} & 46.2                                & 63.6                                 & \bf 65.8                                & \bf 66.1 & 86.1     & 86.4     & 89.7     & 89.9     & 1.68     & 1.89     & 1.26     & 1.24 \\ \cmidrule(lr){1-2}
		\Frag     & \ding{55}         & 45.5                                & 63.7                                 & 64.4                                    & 65.1     & \bf 86.3 & 89.2     & 89.7     & 89.9     & 1.67     & 1.30     & 1.25     & 1.23 \\
		\Frag     & \green{\ding{51}} & \bf 46.7                            & \bf 64.8                             & \bf 65.6                                & \bf 65.6 & \bf 86.5 & \bf 89.6 & \bf 90.0 & \bf 90.1 & 1.59     & 1.24     & 1.20     & 1.20 \\ \cmidrule(lr){1-2}
		\FragDraft   & \ding{55}         & 46.4                                & 64.5                                 & 65.4                                    & 65.3     & 86.4     & 89.3     & 89.8     & 89.8     & 1.65     & 1.28     & 1.23     & 1.23 \\
		\FragDraft   & \green{\ding{51}} & 44.1                                & \bf 65.6                             & \bf 66.3                                & \bf 66.4 & \bf 86.7 & \bf 89.6 & \bf 90.1 & \bf 90.1 & \bf 1.55 & \bf 1.21 & \bf 1.17 & 1.17 \\
		\bottomrule
	\end{tabular}
	\caption{\label{tab:gnome-results} \BLEU, \COMET, and MetricX scores on the GNOME domain,  for several models and number of exemplars $k$. Results significantly better (p<5\%) than the baseline \Baseline are in \textbf{bold}. The significance is assessed with a paired t-test for \COMET and \MetricX, and with bootstrapping for \BLEU (using SacreBLEU, $n=1000$).}
\end{table*}

\subsection{Out-of-domain generalization}

We now challenge the generalization abilities of the reasoning module, evaluating our approach on a domain unseen in fine-tuning: GNOME. Results are in Table~\ref{tab:gnome-results}. The results are similar to those obtained on the in-domain test sets (Table~\ref{tab:main-results}) for the fragment settings, suggesting that fragment-based reasoning is a robust method that can generalize to new domains. Here, the drafting step also yields gains, both with and without fragments.

\subsection{Assessing fragment quality}

Beyond translation scores, a natural follow-up question concerns the validity of extracted fragments: do they correspond to actual chunks matched in the source and the exemplars (RQ1)?
\begin{table}[h]
	\centering
	\begin{tabular}{crrr}
		\toprule
		fragments & Prec. & Rec. & BLEU \\
		\cmidrule(lr){2-4}
		silver    & 98.5  & 81.1 & 77.1 \\
		\Frag         & 99.7  & 89.2 & 80.8 \\
		\FragDraft       & 99.7  & 89.1 & 80.8 \\
		\bottomrule
	\end{tabular}
	\caption{\label{tab:prb-src-vs-frags} Assessing the faithfulness of the source fragments w.r.t. the source sentence.}
\end{table}
To answer this question, we first check whether the source-side fragments are faithful to the source side. Table~\ref{tab:prb-src-vs-frags} reports BLEU scores, unigram recalls and precisions between the concatenated source-side fragments and the source computed with SacreBLEU.

The precision is high, meaning that fragments almost always correspond to spans from the source sentence. The recall is lower, about 80\%, meaning that the source sentence is not represented in its integrality in the fragments. In most cases, this is simply due to ignored punctuation or function words, exactly like in the illustration given in the silver fragment generation prompt (Figure~\ref{fig:qwen-fragment-extraction-prompt}). It is finally noteworthy that the inferred fragments (\Frag and \FragDraft) obtain higher source recall than the silver fragments -- highlighting the positive net effect of SFT for this task.

\subsection{Silver vs. infered fragments and drafts \label{ssec:silver-infered}}

\begin{table*}
	\centering
	\begin{tabular}{lrrrrrrrrrrrr}
		\toprule
		      & \multicolumn{4}{c}{\BLEU$\uparrow$} & \multicolumn{4}{c}{\COMET$\uparrow$} & \multicolumn{4}{c}{MetricX$\downarrow$}                                                                                                \\
		\cmidrule(lr){2-5} \cmidrule(lr){6-9} \cmidrule(lr){10-13}
		Model & k=0                                 & k=1                                  & k=2                                     & k=3      & k=0      & k=1      & k=2      & k=3      & k=0      & k=1      & k=2      & k=3  \\
		\cmidrule(lr){1-1}\cmidrule(lr){2-5} \cmidrule(lr){6-9} \cmidrule(lr){10-13}
		\Draft     & \bf +2.2                            & \bf +1.2                             & \bf +0.7                                & 0.0      & \bf +0.2 & \bf +1.0 & \bf +0.5 & \bf +0.5 & \bf -0.2 & \bf -0.6 & \bf -0.4 & 0.0  \\
		\Frag     & \bf +1.1                            & \bf +0.3                             & \bf +0.1                                & 0.0      & 0.0      & -0.1     & -0.4     & -0.4     & 0.0      & 0.0      & +0.1     & +0.1 \\
		\FragDraft   & \bf +1.3                            & \bf +0.1                             & 0.0                                     & \bf +0.1 & \bf +0.1 & -0.1     & -0.1     & 0.0      & \bf -0.1 & 0.0      & 0.0      & 0.0  \\
		\bottomrule
	\end{tabular}
	\caption{\label{tab:silver-vs-student} \BLEU, \COMET, and MetricX scores average gains on the multi-domain test set when prefixing the silver drafts and/or fragments instead of having them be inferred during reasoning.}
\end{table*}

As the student model attempts to reproduce the silver fragments and/or drafts, errors in the reasoning process may propagate and ultimately degrade the quality of hypotheses. One way to evaluate the quality of the inferred fragments and drafts with respect to the silver ones is to examine the evolution of the translation metrics when the silver fragments and drafts are prefixed to the response as reasoning traces, thereby reducing the generation task to only compute the final translation.
Table~\ref{tab:silver-vs-student} shows that providing the silver fragments alone \Frag does not lead to better translations compared to fully inferring them (except for \BLEU). \COMET scores even hint at a slight negative effect. In contrast, the use of silver drafts yields a clear improvement in all translation scores. These results confirm that the inferred drafts are of comparatively lower quality with respect to the silver drafts.

To further study the difference between silver and generated fragments, we compute their \emph{extraction rate} (ER), which measures how often a fragment is copied into the final translation. ER is the ratio of target-side exact matches between the fragments and the exemplars, conditioned on the source side already being an exact match. Silver fragments obtain an ER of $\approx$70\%, while student fragments (\Frag and \FragDraft) land around 35\%. This loss is fully explained by the fact that the student model sometimes generates fragments instead of extracting them from the exemplars. This does not however impact the overall translation quality.

Finally, we investigate the extent to which the draft constitutes a genuine recombination of fragments, both for the silver traces and the inferred ones (in setting \FragDraft). On average, fragment unigrams cover 81\% of the draft (for silver fragments and drafts), while this proportion rises to 98\% for \FragDraft. These results suggest that silver drafts involve a genuine rephrasing of the target-side fragments, while the student-generated drafts simply amount to recombinations. The poor quality of generated drafts is finally reflected in their BLEU scores: $\approx$ 38 for silver fragments, with generated drafts lagging 8 points behind ($\approx$ 30 in the \FragDraft setting). All this further supports the hypothesis that the drafting stage is unlikely to be beneficial in this setting.

\subsection{Sorting relevant from irrelevant information}

In our scenario, the retrieved exemplars provided to the model are not always close to the source sentence. Some lexical, terminological or phraseological information might be present, but surrounded by less relevant content.
The additional FE step performed during inference should help the identification of these pieces, and their reuse to improve the MT quality. To assess whether this is the case, we study the link between the coverage of the source sentence by the exemplars, and the gain in translation quality between the fragment-based model \Frag and the baseline \Baseline, as measured by \COMET.

\begin{figure}
	\centering
	\includegraphics[width=\linewidth]{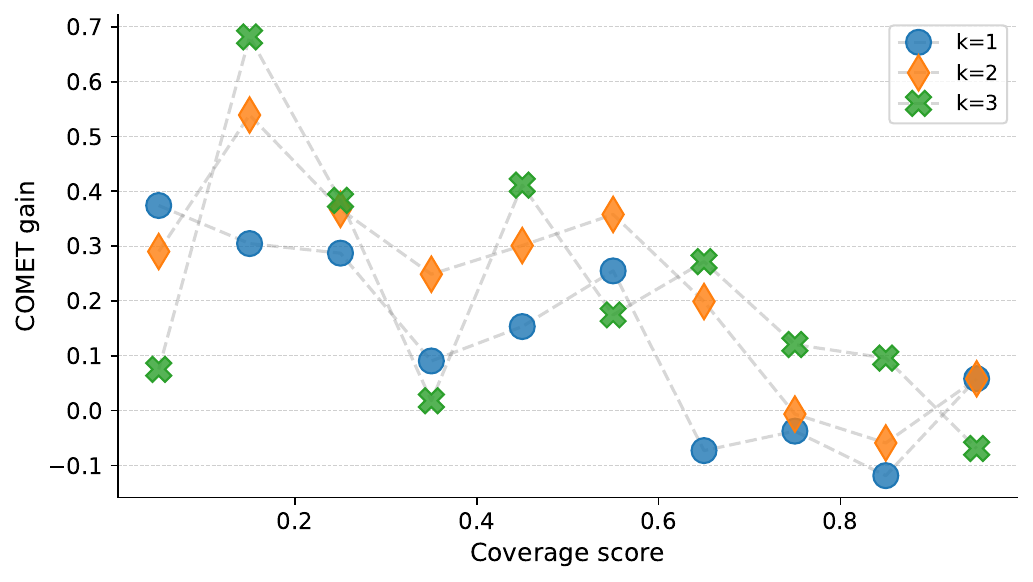}
	\caption{\label{fig:correlation-coverage-comet} Differences in \COMET score gain between \Frag and \Baseline w.r.t the coverage of the reference sentence.}
\end{figure}

The coverage is here defined as the proportion of words in the reference sentence that are also present in the target side of at least one retrieved exemplar.
Figure~\ref{fig:correlation-coverage-comet} reveals that when coverage is low, the \COMET gain is substantially higher than when it  is high, for all values of $k$. This is because a high coverage usually entails a very close exemplar, which by itself ensures a very good translation for the baseline MT. In the low coverage case, where the baseline model can struggle with a lot of irrelevant text, prompting the model to generate fragments has a clear positive effect, as it helps the model to extract relevant information from the exemplars.

\subsection{Domain and language analysis \label{ssec:perdomains}}

Results in Appendix~\ref{appendix:perdomain-scores} display clear and consistent improvements for six domain/language pairs: Wikipedia en-pl, TED en-uk, Europarl en-fr, Wikipedia en-fr, Koran en-de, and Subtitles en-de. French and German generally exhibit stronger gains, probably as they account for 60\% of the training data. In contrast, some domains (e.g., JRC and ECB) and languages (e.g., Spanish and Polish) show more limited improvements. Notably, the majority of the observed gains occur in domains characterized by greater stylistic freedom.

\subsection{Traceability}

Fragments offer a form of traceability of the translation process, as some spans in the hypothesis can be directly linked to spans in the exemplars. Indeed, fragments correspond to an alignment that can be inspected and connected to both the source and the TM exemplars. This traceability feature is illustrated in Appendix~\ref{appendix:traceability}.
However, our model remains free to generate a translation that does not reuse any generated fragments -- and sometimes this is necessary -- making the traceability process sometimes unreliable. Only fragments that are directly copied from the exemplars can be transparently identified.

\fydone{For these analyses, a source of inspiration is \citep{raunak-etal-2023-leveraging}, who proposes various measurements to evaluate the quality of proposed edits.
}
\fytodo{Pour plus tard $k > 3$ : est-ce que là on peut voir briller notre méthode d'extraction face à du k-shot qui stagne ?}
\fytodo{Plus oui moins fait : Qualité des fragments : match/informativité/difficulté}
\fytodo{Mis dans outlook : RL/refinement}
\fytodo{Illustration avec des exemples bons comme mauvais}


\section{Conclusions and Outlook \label{sec:conclusion}}

In this paper, we explore how LLM-based MT can be augmented with a reasoning process that focuses on exemplar matching and recombination. We introduce a fragment-based reasoning framework in which fragments are extracted and/or generated to guide the overall translation process.
We run experiments with six languages and ten domains and observe that our strategy consistently improves translation scores, outperforming standard k-shot prompting and drafting-based approaches. We also noted that the generated fragments mostly correspond to source-side matches; but that the associated draft quality was overall quite low.


Several directions remain open for future work. 
Our current approach relies on supervised distillation from a large teacher model to learn reasoning traces. Recent advances in reasoning-focused LLMs have shown the effectiveness of RL techniques for improving the quality and utility of intermediate reasoning steps, constituting a first promising direction for future research.
Other avenues for future work are the increase of $k$ beyond $3$, as FE seems to robustly handle the additional noise introduced in irrelevant examples. Finally, we would like to better understand the value of the added traceability in user studies involving translators.

\section*{Limitations}

Our work presents several limitations that should be acknowledged. First, although our experiments cover multiple language pairs and a variety of domains, the overall evaluation remains limited in scale. Further validation on a broader and more diverse set of languages, domains, and translation conditions would be necessary to fully assess the generalization and robustness of the proposed approach.

Second, our approach assumes the availability of translation memories (TMs) or large bilingual corpora from which relevant exemplars can be retrieved. In practice, such resources may not always be available, especially for low-resource languages, emerging domains, or highly specialized applications.


Finally, the quality of the proposed reasoning traces strongly depends on the quality of the silver fragments generated by the teacher LLM. In practice, the teacher model may produce suboptimal outputs, including malformed fragments, hallucinated content, incomplete extractions, or incorrect bilingual correspondences. Such errors can propagate to the student models during supervised distillation and ultimately degrade translation quality.
More broadly, the extraction process itself remains constrained by the capabilities and inductive biases of current LLMs. While the generated fragments often capture useful translation correspondences, they do not necessarily reflect linguistically optimal decompositions of the source sentence. Improving the reliability, consistency, and faithfulness of FE therefore remains an important challenge for future work.

\section*{Ethical Statement}
There are no ethical issues with this work.

\section*{Acknowledgments}

This research was funded by the French ``Agence Nationale de la Recherche'' (ANR) under the project TraLaLaM (ANR-23-IAS1-0006). It was provided with computing AI and storage resources by GENCI at IDRIS thanks to grants 2025-A0161015117 and 2026-AD011017822 on the supercomputer Jean Zay.
The authors wish to thank Newman Chen for its contribution to the early stages of this project, and the ARR reviewers and meta-reviewer for their constructive comments.

\bibliography{../anthology,../custom}

\begin{thebibliography}{63}
\providecommand{\natexlab}[1]{#1}

\bibitem[{Agrawal et~al.(2023)Agrawal, Zhou, Lewis, Zettlemoyer, and Ghazvininejad}]{agrawal-etal-2023-context}
Sweta Agrawal, Chunting Zhou, Mike Lewis, Luke Zettlemoyer, and Marjan Ghazvininejad. 2023.
\newblock \href {https://doi.org/10.18653/v1/2023.findings-acl.564} {In-context examples selection for machine translation}.
\newblock In \emph{Findings of the Association for Computational Linguistics: ACL 2023}, pages 8857--8873, Toronto, Canada. Association for Computational Linguistics.

\bibitem[{Aharoni and Goldberg(2020)}]{aharoni-goldberg-2020-unsupervised}
Roee Aharoni and Yoav Goldberg. 2020.
\newblock \href {https://doi.org/10.18653/v1/2020.acl-main.692} {Unsupervised domain clusters in pretrained language models}.
\newblock In \emph{Proceedings of the 58th Annual Meeting of the Association for Computational Linguistics}, pages 7747--7763, Online. Association for Computational Linguistics.

\bibitem[{Bawden and Yvon(2023)}]{bawden-yvon-2023-investigating}
Rachel Bawden and Fran{\c{c}}ois Yvon. 2023.
\newblock \href {https://aclanthology.org/2023.eamt-1.16/} {Investigating the translation performance of a large multilingual language model: the case of {BLOOM}}.
\newblock In \emph{Proceedings of the 24th Annual Conference of the European Association for Machine Translation}, pages 157--170, Tampere, Finland. European Association for Machine Translation.

\bibitem[{Bogoychev et~al.(2023)Bogoychev, van~der Linde, Nail, Haddow, Zaragoza-Bernabeu, Ramírez-Sánchez, Weymann, Mateiu, Helcl, and Aulamo}]{bogoychev-etal-2023-opuscleaner}
Nikolay Bogoychev, Jelmer van~der Linde, Graeme Nail, Barry Haddow, Jaume Zaragoza-Bernabeu, Gema Ramírez-Sánchez, Lukas Weymann, Tudor~Nicolae Mateiu, Jindřich Helcl, and Mikko Aulamo. 2023.
\newblock \href {https://arxiv.org/abs/2311.14838} {Opuscleaner and opustrainer, open source toolkits for training machine translation and large language models}.
\newblock \emph{CoRR}, abs/2311.14838.

\bibitem[{Bouthors et~al.(2023)Bouthors, Crego, and Yvon}]{bouthors-etal-2023-towards}
Maxime Bouthors, Josep Crego, and Fran{\c{c}}ois Yvon. 2023.
\newblock \href {https://doi.org/10.18653/v1/2023.emnlp-main.113} {Towards example-based {NMT} with multi-{L}evenshtein transformers}.
\newblock In \emph{Proceedings of the 2023 Conference on Empirical Methods in Natural Language Processing}, pages 1830--1846, Singapore. Association for Computational Linguistics.

\bibitem[{Bouthors et~al.(2024)Bouthors, Crego, and Yvon}]{bouthors-etal-2024-retrieving}
Maxime Bouthors, Josep Crego, and Fran{\c{c}}ois Yvon. 2024.
\newblock \href {https://doi.org/10.18653/v1/2024.findings-naacl.190} {Retrieving examples from memory for retrieval augmented neural machine translation: A systematic comparison}.
\newblock In \emph{Findings of the Association for Computational Linguistics: NAACL 2024}, pages 3022--3039, Mexico City, Mexico. Association for Computational Linguistics.

\bibitem[{Bowker and Fisher(2010)}]{bowker-fisher-2010}
Lynne Bowker and Desmond Fisher. 2010.
\newblock \href {https://doi.org/10.1075/hts.1.comp2} {\emph{Computer-aided translation}}, pages 60--65.
\newblock Handbook of translation studies. Vol. 1. Amsterdam: Benjamins.

\bibitem[{Briakou et~al.(2024)Briakou, Luo, Cherry, and Freitag}]{briakou-etal-2024-translating}
Eleftheria Briakou, Jiaming Luo, Colin Cherry, and Markus Freitag. 2024.
\newblock \href {https://doi.org/10.18653/v1/2024.wmt-1.123} {Translating step-by-step: Decomposing the translation process for improved translation quality of long-form texts}.
\newblock In \emph{Proceedings of the Ninth Conference on Machine Translation}, pages 1301--1317, Miami, Florida, USA. Association for Computational Linguistics.

\bibitem[{Brown et~al.(2020)Brown, Mann, Ryder, Subbiah, Kaplan, Dhariwal, Neelakantan, Shyam, Sastry, Askell, Agarwal, Herbert-Voss, Krueger, Henighan, Child, Ramesh, Ziegler, Wu, Winter, Hesse, Chen, Sigler, Litwin, Gray, Chess, Clark, Berner, McCandlish, Radford, Sutskever, and Amodei}]{brown-etal-2020-language}
Tom Brown, Benjamin Mann, Nick Ryder, Melanie Subbiah, Jared~D Kaplan, Prafulla Dhariwal, Arvind Neelakantan, Pranav Shyam, Girish Sastry, Amanda Askell, Sandhini Agarwal, Ariel Herbert-Voss, Gretchen Krueger, Tom Henighan, Rewon Child, Aditya Ramesh, Daniel Ziegler, Jeffrey Wu, Clemens Winter, and 12 others. 2020.
\newblock \href {https://proceedings.neurips.cc/paper_files/paper/2020/file/1457c0d6bfcb4967418bfb8ac142f64a-Paper.pdf} {Language models are few-shot learners}.
\newblock In \emph{Advances in Neural Information Processing Systems}, volume~33, pages 1877--1901. Curran Associates, Inc.

\bibitem[{Bulte and Tezcan(2019)}]{bulte-tezcan-2019-neural}
Bram Bulte and Arda Tezcan. 2019.
\newblock \href {https://doi.org/10.18653/v1/P19-1175} {Neural fuzzy repair: Integrating fuzzy matches into neural machine translation}.
\newblock In \emph{Proceedings of the 57th Annual Meeting of the Association for Computational Linguistics}, pages 1800--1809, Florence, Italy. Association for Computational Linguistics.

\bibitem[{Carl et~al.(2004)Carl, Way, and Daelemans}]{carl-etal-2004-recent}
Michael Carl, Andy Way, and Walter Daelemans. 2004.
\newblock \href {https://doi.org/10.1162/0891201042544866} {Recent advances in example-based machine translation}.
\newblock \emph{Computational Linguistics}, 30:516--520.

\bibitem[{Chen et~al.(2025)Chen, Song, Zhu, Chen, Yang, Zhao, and zhang}]{chen-etal-2025-evaluating}
Andong Chen, Yuchen Song, Wenxin Zhu, Kehai Chen, Muyun Yang, Tiejun Zhao, and Min zhang. 2025.
\newblock \href {https://arxiv.org/abs/2502.11544} {Evaluating o1-like {LLMs}: Unlocking reasoning for translation through comprehensive analysis}.
\newblock \emph{Preprint}, arXiv:2502.11544.

\bibitem[{Chen et~al.(2024)Chen, Guo, Haddow, and Heafield}]{chen-etal-2024-iterative}
Pinzhen Chen, Zhicheng Guo, Barry Haddow, and Kenneth Heafield. 2024.
\newblock \href {https://aclanthology.org/2024.eamt-1.17/} {Iterative translation refinement with large language models}.
\newblock In \emph{Proceedings of the 25th Annual Conference of the European Association for Machine Translation (Volume 1)}, pages 181--190, Sheffield, UK. European Association for Machine Translation (EAMT).

\bibitem[{Cheng et~al.(2022)Cheng, Gao, Liu, Zhao, and Yan}]{cheng-etal-2022-neural}
Xin Cheng, Shen Gao, Lemao Liu, Dongyan Zhao, and Rui Yan. 2022.
\newblock \href {https://doi.org/10.18653/v1/2022.emnlp-main.235} {Neural machine translation with contrastive translation memories}.
\newblock In \emph{Proceedings of the 2022 Conference on Empirical Methods in Natural Language Processing}, pages 3591--3601, Abu Dhabi, United Arab Emirates. Association for Computational Linguistics.

\bibitem[{Cobbe et~al.(2021)Cobbe, Kosaraju, Bavarian, Chen, Jun, Kaiser, Plappert, Tworek, Hilton, Nakano, Hesse, and Schulman}]{cobbe-etal-2021-training}
Karl Cobbe, Vineet Kosaraju, Mohammad Bavarian, Mark Chen, Heewoo Jun, Lukasz Kaiser, Matthias Plappert, Jerry Tworek, Jacob Hilton, Reiichiro Nakano, Christopher Hesse, and John Schulman. 2021.
\newblock \href {https://arxiv.org/abs/2110.14168} {Training verifiers to solve math word problems}.
\newblock \emph{CoRR}, abs/2110.14168.

\bibitem[{DeepSeek-AI(2025)}]{deepseekai-2025-deepseekr1}
DeepSeek-AI. 2025.
\newblock \href {https://arxiv.org/abs/2501.12948} {Deepseek-r1: Incentivizing reasoning capability in llms via reinforcement learning}.
\newblock \emph{Preprint}, arXiv:2501.12948.

\bibitem[{DeepSeek-AI et~al.(2025)DeepSeek-AI, Guo, Yang, Zhang, Song, Zhang, Xu, Zhu, Ma, Wang, Bi, Zhang, Yu, Wu, Wu, Gou, Shao, Li, Gao, Liu, Xue, Wang, Wu, Feng, Lu, Zhao, Deng, Zhang, Ruan, Dai, Chen, Ji, Li, Lin, Dai, Luo, Hao, Chen, Li, Zhang, Bao, Xu, Wang, Ding, Xin, Gao, Qu, Li, Guo, Li, Wang, Chen, Yuan, Qiu, Li, Cai, Ni, Liang, Chen, Dong, Hu, Gao, Guan, Huang, Yu, Wang, Zhang, Zhao, Wang, Zhang, Xu, Xia, Zhang, Zhang, Tang, Li, Wang, Li, Tian, Huang, Zhang, Wang, Chen, Du, Ge, Zhang, Pan, Wang, Chen, Jin, Chen, Lu, Zhou, Chen, Ye, Wang, Yu, Zhou, Pan, Li, Zhou, Wu, Ye, Yun, Pei, Sun, Wang, Zeng, Zhao, Liu, Liang, Gao, Yu, Zhang, Xiao, An, Liu, Wang, Chen, Nie, Cheng, Liu, Xie, Liu, Yang, Li, Su, Lin, Li, Jin, Shen, Chen, Sun, Wang, Song, Zhou, Wang, Shan, Li, Wang, Wei, Zhang, Xu, Li, Zhao, Sun, Wang, Yu, Zhang, Shi, Xiong, He, Piao, Wang, Tan, Ma, Liu, Guo, Ou, Wang, Gong, Zou, He, Xiong, Luo, You, Liu, Zhou, Zhu, Xu, Huang, Li, Zheng, Zhu, Ma, Tang, Zha, Yan, Ren, Ren, Sha, Fu, Xu, Xie, Zhang, Hao, Ma, Yan, Wu, Gu, Zhu, Liu, Li, Xie, Song, Pan, Huang, Xu, Zhang, and Zhang}]{deepseekai-etal-2025-deepseek}
DeepSeek-AI, Daya Guo, Dejian Yang, Haowei Zhang, Junxiao Song, Ruoyu Zhang, Runxin Xu, Qihao Zhu, Shirong Ma, Peiyi Wang, Xiao Bi, Xiaokang Zhang, Xingkai Yu, Yu~Wu, Z.~F. Wu, Zhibin Gou, Zhihong Shao, Zhuoshu Li, Ziyi Gao, and 181 others. 2025.
\newblock \href {https://arxiv.org/abs/2501.12948} {Deepseek-r1: Incentivizing reasoning capability in llms via reinforcement learning}.
\newblock \emph{Preprint}, arXiv:2501.12948.

\bibitem[{Feng et~al.(2025)Feng, Cao, Ren, Su, Chen, Zhang, Xu, Hu, Wu, and Liu}]{feng-etal-2025-mtr1}
Zhaopeng Feng, Shaosheng Cao, Jiahan Ren, Jiayuan Su, Ruizhe Chen, Yan Zhang, Zhe Xu, Yao Hu, Jian Wu, and Zuozhu Liu. 2025.
\newblock \href {https://arxiv.org/abs/2504.10160} {{MT-R1-Zero: Advancing LLM-based Machine Translation via R1-Zero-like Reinforcement Learning}}.
\newblock \emph{Preprint}, arXiv:2504.10160.

\bibitem[{Feng et~al.(2024)Feng, Zhang, Li, Wu, Liao, Liu, Lang, Feng, Wu, and Liu}]{feng-etal-2024-tear}
Zhaopeng Feng, Yan Zhang, Hao Li, Bei Wu, Jiayu Liao, Wenqiang Liu, Jun Lang, Yang Feng, Jian Wu, and Zuozhu Liu. 2024.
\newblock \href {https://arxiv.org/abs/2402.16379} {Tear: Improving llm-based machine translation with systematic self-refinement}.
\newblock \emph{Preprint}, arXiv:2402.16379.

\bibitem[{Gu et~al.(2019)Gu, Wang, and Zhao}]{gu-etal-2019-levenshtein}
Jiatao Gu, Changhan Wang, and Junbo Zhao. 2019.
\newblock \href {https://proceedings.neurips.cc/paper/2019/file/675f9820626f5bc0afb47b57890b466e-Paper.pdf} {Levenshtein transformer}.
\newblock In \emph{Advances in Neural Information Processing Systems}, volume~32. Curran Associates, Inc.

\bibitem[{Gu et~al.(2018)Gu, Wang, Cho, and Li}]{gu-etal-2018-search}
Jiatao Gu, Yong Wang, Kyunghyun Cho, and Victor~O.K. Li. 2018.
\newblock \href {https://doi.org/10.1609/aaai.v32i1.12013} {Search {Engine} {Guided} {Neural} {Machine} {Translation}}.
\newblock \emph{Proceedings of the AAAI Conference on Artificial Intelligence}, 32(1).

\bibitem[{He et~al.(2025)He, Liu, Tao, Luo, Zeng, Su, Zhang, Ma, Wei, Meng, Yang, Chen, and Yoshie}]{he-etal-2025-r1t1}
Minggui He, Yilun Liu, Shimin Tao, Yuanchang Luo, Hongyong Zeng, Chang Su, Li~Zhang, Hongxia Ma, Daimeng Wei, Weibin Meng, Hao Yang, Boxing Chen, and Osamu Yoshie. 2025.
\newblock \href {https://arxiv.org/abs/2502.19735} {{R1-T1:} fully incentivizing translation capability in {LLMs} via reasoning learning}.
\newblock \emph{Preprint}, arXiv:2502.19735.

\bibitem[{He et~al.(2021)He, Huang, Cui, Li, and Liu}]{he-etal-2021-fast}
Qiuxiang He, Guoping Huang, Qu~Cui, Li~Li, and Lemao Liu. 2021.
\newblock \href {https://doi.org/10.18653/v1/2021.acl-long.246} {Fast and accurate neural machine translation with translation memory}.
\newblock In \emph{Proceedings of the 59th Annual Meeting of the Association for Computational Linguistics and the 11th International Joint Conference on Natural Language Processing (Volume 1: Long Papers)}, pages 3170--3180, Online. Association for Computational Linguistics.

\bibitem[{He et~al.(2024)He, Liang, Jiao, Zhang, Yang, Wang, Tu, Shi, and Wang}]{he-etal-2024-exploring}
Zhiwei He, Tian Liang, Wenxiang Jiao, Zhuosheng Zhang, Yujiu Yang, Rui Wang, Zhaopeng Tu, Shuming Shi, and Xing Wang. 2024.
\newblock \href {https://doi.org/10.1162/tacl_a_00642} {Exploring human-like translation strategy with large language models}.
\newblock \emph{Transactions of the Association for Computational Linguistics}, 12:229--246.

\bibitem[{Hendy et~al.(2023)Hendy, Abdelrehim, Sharaf, Raunak, Gabr, Matsushita, Kim, Afify, and Awadalla}]{hendy-etal-2023-howgood}
Amr Hendy, Mohamed Abdelrehim, Amr Sharaf, Vikas Raunak, Mohamed Gabr, Hitokazu Matsushita, Young~Jin Kim, Mohamed Afify, and Hany~Hassan Awadalla. 2023.
\newblock \href {https://doi.org/10.48550/ARXIV.2302.09210} {How good are {GPT} models at machine translation? a comprehensive evaluation}.
\newblock \emph{CoRR}, abs/2302.09210.

\bibitem[{Juraska et~al.(2024)Juraska, Deutsch, Finkelstein, and Freitag}]{juraska-etal-2024-metricx}
Juraj Juraska, Daniel Deutsch, Mara Finkelstein, and Markus Freitag. 2024.
\newblock \href {https://doi.org/10.18653/v1/2024.wmt-1.35} {{M}etric{X}-24: The {G}oogle submission to the {WMT} 2024 metrics shared task}.
\newblock In \emph{Proceedings of the Ninth Conference on Machine Translation}, pages 492--504, Miami, Florida, USA. Association for Computational Linguistics.

\bibitem[{Koehn(2010)}]{koehn-2010-smt}
Philipp Koehn. 2010.
\newblock \emph{Statistical machine translation}.
\newblock Cambridge University Press.

\bibitem[{Kojima et~al.(2022)Kojima, Gu, Reid, Matsuo, and Iwasawa}]{kojima-etal-2022-large}
Takeshi Kojima, Shixiang~(Shane) Gu, Machel Reid, Yutaka Matsuo, and Yusuke Iwasawa. 2022.
\newblock Large language models are zero-shot reasoners.
\newblock In \emph{Advances in Neural Information Processing Systems}, volume~35, pages 22199--22213.

\bibitem[{Lommel et~al.(2013)Lommel, Burchardt, and Uszkoreit}]{burchardt-2013-multidimensional}
Arle~Richard Lommel, Aljoscha Burchardt, and Hans Uszkoreit. 2013.
\newblock \href {https://aclanthology.org/2013.tc-1.6/} {Multidimensional quality metrics: a flexible system for assessing translation quality}.
\newblock In \emph{Proceedings of Translating and the Computer 35}, London, UK. Aslib.

\bibitem[{Moslem et~al.(2023)Moslem, Haque, Kelleher, and Way}]{moslem-etal-2023-adaptive}
Yasmin Moslem, Rejwanul Haque, John~D. Kelleher, and Andy Way. 2023.
\newblock \href {https://aclanthology.org/2023.eamt-1.22/} {Adaptive machine translation with large language models}.
\newblock In \emph{Proceedings of the 24th Annual Conference of the European Association for Machine Translation}, pages 227--237, Tampere, Finland. European Association for Machine Translation.

\bibitem[{Nagao(1984)}]{nagao-1984-framework}
Makoto Nagao. 1984.
\newblock \href {https://mt-archive.net/70/Nagao-1984.pdf} {A framework of a mechanical translation between {Japanese} and {English} by analogy principle}.
\newblock In \emph{Artificial and human intelligence}. Elsevier Science Publishers. B.V.

\bibitem[{Nguyen and Xu(2025)}]{nguyen-xu-2025-reasoning}
Lam Nguyen and Yang Xu. 2025.
\newblock \href {https://doi.org/10.18653/v1/2025.acl-srw.17} {Reasoning for translation: Comparative analysis of chain-of-thought and tree-of-thought prompting for {LLM} translation}.
\newblock In \emph{Proceedings of the 63rd Annual Meeting of the Association for Computational Linguistics (Volume 4: Student Research Workshop)}, pages 259--275, Vienna, Austria. Association for Computational Linguistics.

\bibitem[{Niwa et~al.(2022)Niwa, Takase, and Okazaki}]{niwa-etal-2022-nearest}
Ayana Niwa, Sho Takase, and Naoaki Okazaki. 2022.
\newblock \href {https://doi.org/10.48550/ARXIV.2208.12496} {Nearest neighbor non-autoregressive text generation}.
\newblock \emph{CoRR}, abs/2208.12496.

\bibitem[{Papineni et~al.(2002)Papineni, Roukos, Ward, and Zhu}]{papineni-etal-2002-bleu}
Kishore Papineni, Salim Roukos, Todd Ward, and Wei-Jing Zhu. 2002.
\newblock \href {https://doi.org/10.3115/1073083.1073135} {{B}leu: a method for automatic evaluation of machine translation}.
\newblock In \emph{Proceedings of the 40th Annual Meeting of the Association for Computational Linguistics}, pages 311--318, Philadelphia, Pennsylvania, USA. Association for Computational Linguistics.

\bibitem[{Post(2018)}]{post-2018-call}
Matt Post. 2018.
\newblock \href {https://doi.org/10.18653/v1/W18-6319} {A call for clarity in reporting {BLEU} scores}.
\newblock In \emph{Proceedings of the Third Conference on Machine Translation: Research Papers}, pages 186--191, Brussels, Belgium. Association for Computational Linguistics.

\bibitem[{Proietti et~al.(2025)Proietti, Perrella, Zouhar, Navigli, and Kocmi}]{proietti-etal-2025-estimating}
Lorenzo Proietti, Stefano Perrella, Vil{\'e}m Zouhar, Roberto Navigli, and Tom Kocmi. 2025.
\newblock \href {https://doi.org/10.18653/v1/2025.findings-emnlp.1317} {Estimating machine translation difficulty}.
\newblock In \emph{Findings of the Association for Computational Linguistics: EMNLP 2025}, pages 24261--24285, Suzhou, China. Association for Computational Linguistics.

\bibitem[{Radford et~al.(2019)Radford, Wu, Child, Luan, Amodei, Sutskever et~al.}]{radford-etal-2019-language}
Alec Radford, Jeffrey Wu, Rewon Child, David Luan, Dario Amodei, Ilya Sutskever, and 1 others. 2019.
\newblock Language models are unsupervised multitask learners.
\newblock \emph{OpenAI blog}, 1(8):9.

\bibitem[{Rajaee et~al.(2026)Rajaee, Vincent, Berard, Fadaee, Marchisio, and Kocmi}]{rajaee-etal-2026-unlocking}
Sara Rajaee, Sebastian Vincent, Alexandre Berard, Marzieh Fadaee, Kelly Marchisio, and Tom Kocmi. 2026.
\newblock \href {https://arxiv.org/abs/2602.14763} {Unlocking reasoning capability on machine translation in large language models}.
\newblock \emph{Preprint}, arXiv:2602.14763.

\bibitem[{Raunak et~al.(2023)Raunak, Sharaf, Wang, Awadalla, and Menezes}]{raunak-etal-2023-leveraging}
Vikas Raunak, Amr Sharaf, Yiren Wang, Hany Awadalla, and Arul Menezes. 2023.
\newblock \href {https://doi.org/10.18653/v1/2023.findings-emnlp.804} {Leveraging {GPT}-4 for automatic translation post-editing}.
\newblock In \emph{Findings of the Association for Computational Linguistics: EMNLP 2023}, pages 12009--12024, Singapore. Association for Computational Linguistics.

\bibitem[{Rei et~al.(2020)Rei, Stewart, Farinha, and Lavie}]{rei-etal-2020-comet}
Ricardo Rei, Craig Stewart, Ana~C Farinha, and Alon Lavie. 2020.
\newblock \href {https://doi.org/10.18653/v1/2020.emnlp-main.213} {{COMET}: A neural framework for {MT} evaluation}.
\newblock In \emph{Proceedings of the 2020 Conference on Empirical Methods in Natural Language Processing (EMNLP)}, pages 2685--2702, Online. Association for Computational Linguistics.

\bibitem[{Rudin(2019)}]{rudin-cynthia-2019-stop}
Cynthia Rudin. 2019.
\newblock \href {https://doi.org/10.1038/s42256-019-0048-x} {Stop explaining black box machine learning models for high stakes decisions and use interpretable models instead}.
\newblock \emph{Nature Machine Intelligence}, 1(5):206--215.

\bibitem[{Setlur et~al.(2024)Setlur, Nagpal, Fisch, Geng, Eisenstein, Agarwal, Agarwal, Berant, and Kumar}]{setlur-etal-2024-rewarding}
Amrith Setlur, Chirag Nagpal, Adam Fisch, Xinyang Geng, Jacob Eisenstein, Rishabh Agarwal, Alekh Agarwal, Jonathan Berant, and Aviral Kumar. 2024.
\newblock \href {https://arxiv.org/abs/2410.08146} {Rewarding progress: Scaling automated process verifiers for llm reasoning}.
\newblock \emph{Preprint}, arXiv:2410.08146.

\bibitem[{Shum et~al.(2023)Shum, Diao, and Zhang}]{shum-etal-2023-automatic}
Kashun Shum, Shizhe Diao, and Tong Zhang. 2023.
\newblock \href {https://doi.org/10.18653/v1/2023.findings-emnlp.811} {Automatic prompt augmentation and selection with chain-of-thought from labeled data}.
\newblock In \emph{Findings of the Association for Computational Linguistics: EMNLP 2023}, pages 12113--12139, Singapore. Association for Computational Linguistics.

\bibitem[{Somers(1999)}]{somers-1999-review}
Harold Somers. 1999.
\newblock \href {https://doi.org/10.1023/A:1008109312730} {Review article: Example-based machine translation}.
\newblock \emph{Machine Translation}, 14(2):113--157.

\bibitem[{Tiedemann(2012)}]{tiedemann-2012-parallel}
J{\"o}rg Tiedemann. 2012.
\newblock \href {https://aclanthology.org/L12-1246/} {Parallel data, tools and interfaces in {OPUS}}.
\newblock In \emph{Proceedings of the Eighth International Conference on Language Resources and Evaluation ({LREC}'12)}, pages 2214--2218, Istanbul, Turkey. European Language Resources Association (ELRA).

\bibitem[{Touvron et~al.(2023)Touvron, Martin, Stone, Albert, Almahairi, Babaei, Bashlykov, Batra, Bhargava, Bhosale, Bikel, Blecher, Ferrer, Chen, Cucurull, Esiobu, Fernandes, Fu, Fu, Fuller, Gao, Goswami, Goyal, Hartshorn, Hosseini, Hou, Inan, Kardas, Kerkez, Khabsa, Kloumann, Korenev, Koura, Lachaux, Lavril, Lee, Liskovich, Lu, Mao, Martinet, Mihaylov, Mishra, Molybog, Nie, Poulton, Reizenstein, Rungta, Saladi, Schelten, Silva, Smith, Subramanian, Tan, Tang, Taylor, Williams, Kuan, Xu, Yan, Zarov, Zhang, Fan, Kambadur, Narang, Rodriguez, Stojnic, Edunov, and Scialom}]{touvron-etal-2023-llama}
Hugo Touvron, Louis Martin, Kevin Stone, Peter Albert, Amjad Almahairi, Yasmine Babaei, Nikolay Bashlykov, Soumya Batra, Prajjwal Bhargava, Shruti Bhosale, Dan Bikel, Lukas Blecher, Cristian~Canton Ferrer, Moya Chen, Guillem Cucurull, David Esiobu, Jude Fernandes, Jeremy Fu, Wenyin Fu, and 49 others. 2023.
\newblock \href {https://arxiv.org/abs/2307.09288} {Llama 2: Open foundation and fine-tuned chat models}.
\newblock \emph{CoRR}, abs/2307.09288.

\bibitem[{Vilar et~al.(2023)Vilar, Freitag, Cherry, Luo, Ratnakar, and Foster}]{vilar-etal-2023-prompting}
David Vilar, Markus Freitag, Colin Cherry, Jiaming Luo, Viresh Ratnakar, and George Foster. 2023.
\newblock \href {https://doi.org/10.18653/v1/2023.acl-long.859} {Prompting {P}a{LM} for translation: Assessing strategies and performance}.
\newblock In \emph{Proceedings of the 61st Annual Meeting of the Association for Computational Linguistics (Volume 1: Long Papers)}, pages 15406--15427, Toronto, Canada. Association for Computational Linguistics.

\bibitem[{Vilar et~al.(2022)Vilar, Freitag, Cherry, Luo, Ratnakar, and Foster}]{vilar-etal-2022-prompting}
David Vilar, Markus Freitag, Colin Cherry, Jiaming Luo, Viresh Ratnakar, and George~F. Foster. 2022.
\newblock \href {https://doi.org/10.48550/arXiv.2211.09102} {{Prompting PaLM for Translation: Assessing Strategies and Performance}}.
\newblock \emph{CoRR}, abs/2211.09102.

\bibitem[{Wei et~al.(2022)Wei, Wang, Schuurmans, Bosma, ichter, Xia, Chi, Le, and Zhou}]{wei-etal-2022-chain}
Jason Wei, Xuezhi Wang, Dale Schuurmans, Maarten Bosma, brian ichter, Fei Xia, Ed~Chi, Quoc~V Le, and Denny Zhou. 2022.
\newblock \href {https://proceedings.neurips.cc/paper_files/paper/2022/file/9d5609613524ecf4f15af0f7b31abca4-Paper-Conference.pdf} {Chain-of-thought prompting elicits reasoning in large language models}.
\newblock In \emph{Advances in Neural Information Processing Systems}, volume~35, pages 24824--24837. Curran Associates, Inc.

\bibitem[{Xia et~al.(2019)Xia, Huang, Liu, and Shi}]{xia-etal-2019-graph}
Mengzhou Xia, Guoping Huang, Lemao Liu, and Shuming Shi. 2019.
\newblock \href {https://doi.org/10.1609/aaai.v33i01.33017297} {{Graph Based Translation-Memory for Neural Machine Translation}}.
\newblock In \emph{Proceedings of the AAAI Conference on Artificial Intelligence}, AAAAI, pages 7297--7304.

\bibitem[{Xu et~al.(2022)Xu, Crego, and Senellart}]{xu-etal-2022-boosting}
Jitao Xu, Josep Crego, and Jean Senellart. 2022.
\newblock \href {https://aclanthology.org/2022.amta-upg.20/} {Boosting neural machine translation with similar translations}.
\newblock In \emph{Proceedings of the 15th Biennial Conference of the Association for Machine Translation in the Americas (Volume 2: Users and Providers Track and Government Track)}, pages 282--292, Orlando, USA. Association for Machine Translation in the Americas.

\bibitem[{Xu et~al.(2023)Xu, Crego, and Yvon}]{xu-etal-2023-integrating}
Jitao Xu, Josep Crego, and Fran{\c{c}}ois Yvon. 2023.
\newblock \href {https://doi.org/10.18653/v1/2023.eacl-main.96} {Integrating translation memories into non-autoregressive machine translation}.
\newblock In \emph{Proceedings of the 17th Conference of the European Chapter of the Association for Computational Linguistics}, pages 1326--1338, Dubrovnik, Croatia. Association for Computational Linguistics.

\bibitem[{Xu and Carpuat(2021)}]{xu_editor_2021}
Weijia Xu and Marine Carpuat. 2021.
\newblock \href {https://doi.org/10.1162/tacl_a_00368} {{EDITOR}: {An} {Edit}-{Based} {Transformer} with {Repositioning} for {Neural} {Machine} {Translation} with {Soft} {Lexical} {Constraints}}.
\newblock \emph{Transactions of the Association for Computational Linguistics}, 9:311--328.

\bibitem[{Xu et~al.(2024)Xu, Deutsch, Finkelstein, Juraska, Zhang, Liu, Wang, Li, and Freitag}]{xu-etal-2024-llmrefine}
Wenda Xu, Daniel Deutsch, Mara Finkelstein, Juraj Juraska, Biao Zhang, Zhongtao Liu, William~Yang Wang, Lei Li, and Markus Freitag. 2024.
\newblock \href {https://doi.org/10.18653/v1/2024.findings-naacl.92} {{LLMR}efine: Pinpointing and refining large language models via fine-grained actionable feedback}.
\newblock In \emph{Findings of the Association for Computational Linguistics: NAACL 2024}, pages 1429--1445, Mexico City, Mexico. Association for Computational Linguistics.

\bibitem[{Yang et~al.(2025)Yang, Li, Yang, Zhang, Hui, Zheng, Yu, Gao, Huang, Lv, Zheng, Liu, Zhou, Huang, Hu, Ge, Wei, Lin, Tang, Yang, Tu, Zhang, Yang, Yang, Zhou, Zhou, Lin, Dang, Bao, Yang, Yu, Deng, Li, Xue, Li, Zhang, Wang, Zhu, Men, Gao, Liu, Luo, Li, Tang, Yin, Ren, Wang, Zhang, Ren, Fan, Su, Zhang, Zhang, Wan, Liu, Wang, Cui, Zhang, Zhou, and Qiu}]{yang-etal-2025-qwen3}
An~Yang, Anfeng Li, Baosong Yang, Beichen Zhang, Binyuan Hui, Bo~Zheng, Bowen Yu, Chang Gao, Chengen Huang, Chenxu Lv, Chujie Zheng, Dayiheng Liu, Fan Zhou, Fei Huang, Feng Hu, Hao Ge, Haoran Wei, Huan Lin, Jialong Tang, and 41 others. 2025.
\newblock \href {https://arxiv.org/abs/2505.09388} {Qwen3 technical report}.
\newblock \emph{Preprint}, arXiv:2505.09388.

\bibitem[{Yasunaga et~al.(2024)Yasunaga, Chen, Li, Pasupat, Leskovec, Liang, Chi, and Zhou}]{yasunaga-etal-2024-large}
Michihiro Yasunaga, Xinyun Chen, Yujia Li, Panupong Pasupat, Jure Leskovec, Percy Liang, Ed~H. Chi, and Denny Zhou. 2024.
\newblock \href {https://openreview.net/forum?id=AgDICX1h50} {Large language models as analogical reasoners}.
\newblock In \emph{The Twelfth International Conference on Learning Representations}.

\bibitem[{Zebaze et~al.(2025{\natexlab{a}})Zebaze, Bawden, and Sagot}]{zebaze-etal-2025-reasoning}
Armel Zebaze, Rachel Bawden, and Benoît Sagot. 2025{\natexlab{a}}.
\newblock \href {https://arxiv.org/abs/2510.11919} {{LLM} reasoning for machine translation: Synthetic data generation over thinking tokens}.
\newblock \emph{Preprint}, arXiv:2510.11919.

\bibitem[{Zebaze et~al.(2025{\natexlab{b}})Zebaze, Sagot, and Bawden}]{zebaze-etal-2025-compositional}
Armel~Randy Zebaze, Beno{\^i}t Sagot, and Rachel Bawden. 2025{\natexlab{b}}.
\newblock \href {https://doi.org/10.18653/v1/2025.findings-emnlp.1216} {Compositional translation: A novel {LLM}-based approach for low-resource machine translation}.
\newblock In \emph{Findings of the Association for Computational Linguistics: EMNLP 2025}, pages 22328--22357, Suzhou, China. Association for Computational Linguistics.

\bibitem[{Zebaze et~al.(2025{\natexlab{c}})Zebaze, Sagot, and Bawden}]{zebaze-etal-2025-context}
Armel~Randy Zebaze, Beno{\^i}t Sagot, and Rachel Bawden. 2025{\natexlab{c}}.
\newblock \href {https://doi.org/10.18653/v1/2025.findings-naacl.68} {In-context example selection via similarity search improves low-resource machine translation}.
\newblock In \emph{Findings of the Association for Computational Linguistics: NAACL 2025}, pages 1222--1252, Albuquerque, New Mexico. Association for Computational Linguistics.

\bibitem[{Zhang et~al.(2023{\natexlab{a}})Zhang, Haddow, and Birch}]{zhang-etal-2023-prompting}
Biao Zhang, Barry Haddow, and Alexandra Birch. 2023{\natexlab{a}}.
\newblock \href {https://proceedings.mlr.press/v202/zhang23m.html} {Prompting large language model for machine translation: A case study}.
\newblock In \emph{Proceedings of the 40th International Conference on Machine Learning}, ICML'23. JMLR.org.

\bibitem[{Zhang et~al.(2022)Zhang, Roller, Goyal, Artetxe, Chen, Chen, Dewan, Diab, Li, Lin, Mihaylov, Ott, Shleifer, Shuster, Simig, Koura, Sridhar, Wang, and Zettlemoyer}]{zhang-etal-2022-opt}
Susan Zhang, Stephen Roller, Naman Goyal, Mikel Artetxe, Moya Chen, Shuohui Chen, Christopher Dewan, Mona Diab, Xian Li, Xi~Victoria Lin, Todor Mihaylov, Myle Ott, Sam Shleifer, Kurt Shuster, Daniel Simig, Punit~Singh Koura, Anjali Sridhar, Tianlu Wang, and Luke Zettlemoyer. 2022.
\newblock \href {https://arxiv.org/abs/2205.01068} {Opt: Open pre-trained transformer language models}.
\newblock \emph{CoRR}, abs/2205.01068.

\bibitem[{Zhang et~al.(2023{\natexlab{b}})Zhang, Zhang, Li, and Smola}]{zhang-etal-2023-automatic}
Zhuosheng Zhang, Aston Zhang, Mu~Li, and Alex Smola. 2023{\natexlab{b}}.
\newblock \href {https://openreview.net/forum?id=5NTt8GFjUHkr} {Automatic chain of thought prompting in large language models}.
\newblock In \emph{The Eleventh International Conference on Learning Representations}.

\bibitem[{Zheng et~al.(2023)Zheng, Wang, Wang, Chen, Zhang, and Tu}]{zheng-etal-2023-towards}
Kangjie Zheng, Longyue Wang, Zhihao Wang, Binqi Chen, Ming Zhang, and Zhaopeng Tu. 2023.
\newblock \href {https://doi.org/10.1109/ICASSP49357.2023.10094646} {Towards a unified training for {Levenshtein} transformer}.
\newblock In \emph{Proceedings of the IEEE International Conference on Acoustics, Speech and Signal Processing (ICASSP)}, pages 1--5.

\end{thebibliography}

\appendix

\section{Datasets \label{appendix:data}}

\begin{table*}
	\centering
	\small
	\begin{tabular}{crrrrrrrrrrrrrrrr}
		\toprule
		domain                              & \makebox[\widthof{xxx}]{ECB} & \makebox[\widthof{xxx}]{JRC} & \makebox[\widthof{xxx}]{Wiki} & \makebox[\widthof{xxx}]{JRC}  & \makebox[\widthof{xxx}]{KDE} & \makebox[\widthof{xxx}]{TED}  & \makebox[\widthof{xxx}]{ECB}  & \makebox[\widthof{xxx}]{EMEA} & \makebox[\widthof{xxx}]{Epp}  & \makebox[\widthof{xxx}]{JRC}  & \makebox[\widthof{xxx}]{Wiki} & \makebox[\widthof{xxx}]{KDE}  & \makebox[\widthof{xxx}]{EMEA} & \makebox[\widthof{xxx}]{JRC}  & \makebox[\widthof{xxx}]{Kor}  & \makebox[\widthof{xxx}]{Sub}  \\ \cmidrule(lr){2-17}
		\makebox[\widthof{xxx}]{en-xx pair} & pl                           & pl                           & pl                            & es                            & uk                           & uk                            & fr                            & fr                            & fr                            & fr                            & fr                            & de                            & de                            & de                            & de                            & de                            \\ \cmidrule(lr){2-17}
		size                                & 46K                          & \makebox[\widthof{xxx}]{1.0M}                           & \makebox[\widthof{xxx}]{111K} & \makebox[\widthof{xxx}]{480K} & 93K                          & \makebox[\widthof{xxx}]{177K} & \makebox[\widthof{xxx}]{140K} & \makebox[\widthof{xxx}]{242K} & \makebox[\widthof{xxx}]{1.9M} & \makebox[\widthof{xxx}]{475K} & \makebox[\widthof{xxx}]{538K} & \makebox[\widthof{xxx}]{223K} & \makebox[\widthof{xxx}]{18K}  & \makebox[\widthof{xxx}]{467K} & \makebox[\widthof{xxx}]{248K} & \makebox[\widthof{xxx}]{500K} \\
		\bottomrule
	\end{tabular}
	\caption{\label{tab:data-stats}List of domains with the size of the full available training datasets.}
\end{table*}

The corpus comprises various domains:
\begin{itemize}
	\item IT (KDE)
	\item Legal (JRC)
	\item Finance (ECB)
	\item Medical (EMEA)
	\item Religion (Kor)
	\item Politics (Europarl)
	\item TED talks
	\item Wikipedia
	\item Subtitles
\end{itemize}

The French-English and German-English datasets are respectively from \citet{bouthors-etal-2024-retrieving} and \citet{aharoni-goldberg-2020-unsupervised}, and are already filtered and split into train/dev/test.
The other datasets, are downloaded from OPUS \citep{tiedemann-2012-parallel} with OpusCleaner\footnote{\url{https://github.com/hplt-project/OpusCleaner}} \citep{bogoychev-etal-2023-opuscleaner} in order to remove noisy samples: empty sentences and too short/long sentences (4-150 words).
In a second step, we compute \COMETKiwi scores \citep{rei-etal-2020-comet} for all the parallel sentences in order to filter out those with a score below a threshold computed automatically.\footnote{\url{https://github.com/Maxwell1447/Auto-Threshold-For-Data-Filtering}}

The remaining data is split into train/dev/test sets when not already available (like for the en-de domains of \citet{aharoni-goldberg-2020-unsupervised} or the en-fr domains of \citep{bouthors-etal-2024-retrieving}). Our custom dev sets each contain 100 samples, while test sets contain 1000 samples.

In order to select challenging high quality data for training, we select a subset of 10K samples for each domain that obtain the highest "quality" scores. This quality score is an aggregation of the \COMETKiwi score and \sentinelsrc score\footnote{\url{https://huggingface.co/Prosho/sentinel-src-25}} \citep{proietti-etal-2025-estimating}. \COMETKiwi estimates the parallel quality of a sentence pair, while \sentinelsrc estimates the difficulty of translating a given source sentence. The quality score is computed as:
\begin{equation}
	\text{Quality}(\src, \tgt) = \text{QE}(\src, \tgt)^2 \cdot \text{DIFF}(\src);
\end{equation}
where $\text{QE}(\src, \tgt)$ is the \COMETKiwi score of the sentence pair and $\text{DIFF}(\src)$ is one minus the \sentinelsrc score of the source sentence. The \COMETKiwi score is squared in order to give more weight to parallelism, in order to avoid having difficult low quality samples in the training data.

For each domain, we select a subset of 10k samples for training, and concatenate these subsets into a multilingual dataset of size 160k.
Eventually, we perform a retrieval step to obtain similar exemplars for each of the training, validation and test samples. For each instance, we uniformly sample between $k=0$ and $k=3$ exemplars by retrieving in-domain similar sentences from the pool of all parallel sentences (excluding dev and test splits). The retrieval setting leverages BM25 and the Levenshtein distance, as advised by \citet{bouthors-etal-2024-retrieving} using an open-source fuzzy-matching tool\footnote{\url{https://github.com/SYSTRAN/fuzzy-match}}.

\section{Fine-tuning of Qwen-8B for fragment-based MT\label{appendix:qwen-ft}}

The supervised fine-tuning of the Qwen3-8B models is performed with HuggingFace. The setting uses LoRA with rank 16, $\alpha=16$, and a dropout of 0.05 on layers Q, K, V, O, Gates, Up and Down. The training lasts for exactly two epochs (according to preliminary results which suggest that it is the optimal number of epochs), with a learning rate of 2e-4 and a cosine scheduler, a warmup ratio of 0.05, AdamW as the optimizer with a weight decay of 0.01, and an effective batch size of 32.

\section{Prompting Qwen to extract silver fragments\label{appendix:qwen-fragments-prompting}}

Figure~\ref{fig:qwen-fragment-extraction-prompt} presents the FE prompt used to extract both the silver fragments and the draft. Notably, the prompt does not incorporate the reference translation. This design choice is motivated by preliminary experiments indicating that, when the reference is provided, the target-side fragments tend to be copied directly from it rather than extracted from the exemplars. Furthermore, in such cases, there is no guarantee that the generated draft is genuinely a draft rather than a simple reproduction of the reference. This behavior is mitigated by completely withholding the reference from the teacher model.

\section{Prompting Qwen-8B for fragment-based MT\label{appendix:qwen-translation-prompting}}

The following figures (\ref{fig:qwen-ft-baseline}, \ref{fig:qwen-ft-draft}, \ref{fig:qwen-ft-fragments}, \ref{fig:qwen-ft-draft-fragments}) illustrate the prompts used to train the different student models.
All prompts use the chatML format, already used by the Qwen3 model family.
The prompt used for the baseline model \Baseline is quite simple, as it only requires to generate the translation without any extra information.
Note that in every prompts, only the \emph{thinking} changes. The expected reasoning format is induced by the the training data during SFT, since there is one model for each prompt. Notably, around 20\% of the prompts given to the \Draft, \Frag, and \FragDraft models during training contain empty reasoning responses like in the baseline \Baseline.

\section{Detailed results per domain \label{appendix:perdomain-scores}}

We mostly presented aggregated results across all 16 domains/languages, even though the scores and gains may vary depending on them. The full picture of \BLEU, \COMET and \MetricX scores for each model (\Instruct, \Baseline, \Draft, \Frag and \FragDraft) are displayed repectively in tables~\ref{tab:model-I}, \ref{tab:model-B}, \ref{tab:model-D}, \ref{tab:model-F} and \ref{tab:model-FD}.

\begin{figure*}
	\centering
	\begin{tcolorbox}[
			title=Silver Fragment \& Draft Generation Prompt,
			coltitle=white,
			colbacktitle=blue!60!green,
			fontupper=\ttfamily\small
		]

		\begin{tcolorbox}[
				colback=blue!5,
				colframe=blue!60!green,
				boxrule=0.8pt,
				title=User,
				sharp corners,
				left=2mm,right=2mm,top=2mm,bottom=2mm
			]
			Given a source sentence, and a potential list of similar examples:\\
			* Identify the translation of the spans constituting the source from the examples\\
			* Draft a candidate translation of the source sentence\\

			\# STEP 1: SPANS\\

			First, extract a list of small parallel spans from the source sentence whose translation may be present in the examples in the target language. Spans should be small semantic units.\\

			\# STEP 2: DRAFTING\\

			Combine the useful spans to form a candidate translation of the source sentence.\\

			\# ILLUSTRATION\\

			The output must comply with this illustration:\\
			Source: For the next years, the amount of such standardised deductions will be published by the ECB.\\

			Example 1:\\
			English: For the next weeks, the Commission shall draw up a list of the agents.\\
			French: Pour les semaines suivantes, la Commission établit la liste des agents.\\

			Example 2:\\
			English: The standard deduction shall be published by the ECB in the same manner as the publication of the list referred to in Article 2(3).\\
			French: La déduction forfaitaire est publié par la BCE de la même manière que la liste mentionnée à l'article 2, paragraphe 3.\\

			The output should follow this format:\\
			<OUT>\\
			*   Extracted spans:\\
			\indent\hspace{1cm} *   For the next years -> Pour les années suivantes\\
			\indent\hspace{1cm} *   the amount of -> le montant de\\
			\indent\hspace{1cm} *   standardised deductions -> déductions forfaitaires\\
			\indent\hspace{1cm} *   will be published -> sera publié\\
			\indent\hspace{1cm} *   by the ECB -> par la BCE\\
			*   Drafting:\\
			Pour les années suivantes, le montant de ces déductions forfaitaires sera publié par la BCE.\\
			</OUT>\\

			\# CONTENT\\
			Here are the source sentence along with the similar \{src\_lng\}-\{tgt\_lng\} examples:\\

			Source: \{source\} \\

			Example 1:\\
			\{src\_lng\}: \{src\_example\_1\} \\
			\{tgt\_lng\}: \{tgt\_example\_1\} \\
			$\left[...\right]$\\
			Example k:\\
			\{src\_lng\}: \{src\_example\_k\} \\
			\{tgt\_lng\}: \{tgt\_example\_k\} \\

		\end{tcolorbox}


	\end{tcolorbox}

	\caption{\label{fig:qwen-fragment-extraction-prompt} FE prompt used to generate the silver fragments an the draft with \texttt{Qwen3-32B}. Thinking is disabled.}
\end{figure*}

\begin{figure*}
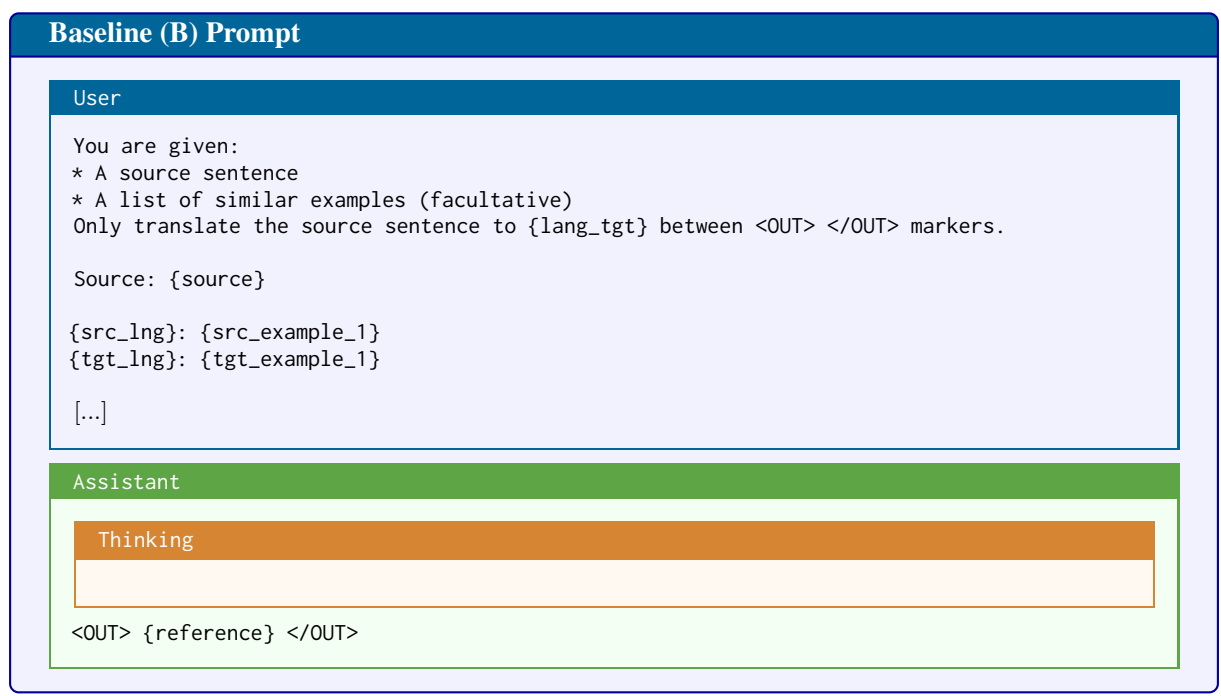

	\centering
	\begin{tcolorbox}[
			title=Baseline (B) Prompt,
			coltitle=white,
			colbacktitle=blue!60!green,
			fontupper=\ttfamily\small
		]

		\begin{tcolorbox}[
				colback=blue!5,
				colframe=blue!60!green,
				boxrule=0.8pt,
				title=User,
				sharp corners,
				left=2mm,right=2mm,top=2mm,bottom=2mm
			]
			You are given:\\
			* A source sentence\\
			* A list of similar examples (facultative)\\
			Only translate the source sentence to \{lang\_tgt\} between <OUT> </OUT> markers.\\

			Source: \{source\} \\

			\{src\_lng\}: \{src\_example\_1\} \\
			\{tgt\_lng\}: \{tgt\_example\_1\} \\

			$\left[...\right]$

		\end{tcolorbox}

		\begin{tcolorbox}[
				colback=green!5,
				colframe=green!60!black!70!white!90!red,
				boxrule=0.8pt,
				sharp corners,
				title=Assistant,
				left=2mm,right=2mm,top=2mm,bottom=2mm
			]
			\begin{tcolorbox}[
					colback=orange!5,
					colframe=orange!80!black!80!white,
					boxrule=0.8pt,
					sharp corners,
					title=Thinking,
					left=2mm,right=2mm,top=2mm,bottom=2mm
				]
			\end{tcolorbox}
			<OUT>
			\{reference\}
			</OUT>

		\end{tcolorbox}

	\end{tcolorbox}

	\caption{\label{fig:qwen-ft-baseline} Prompt used to train the baseline model \Baseline. The thinking process is empty, as the model is only trained to generate a translation.}
\end{figure*}

\begin{figure*}
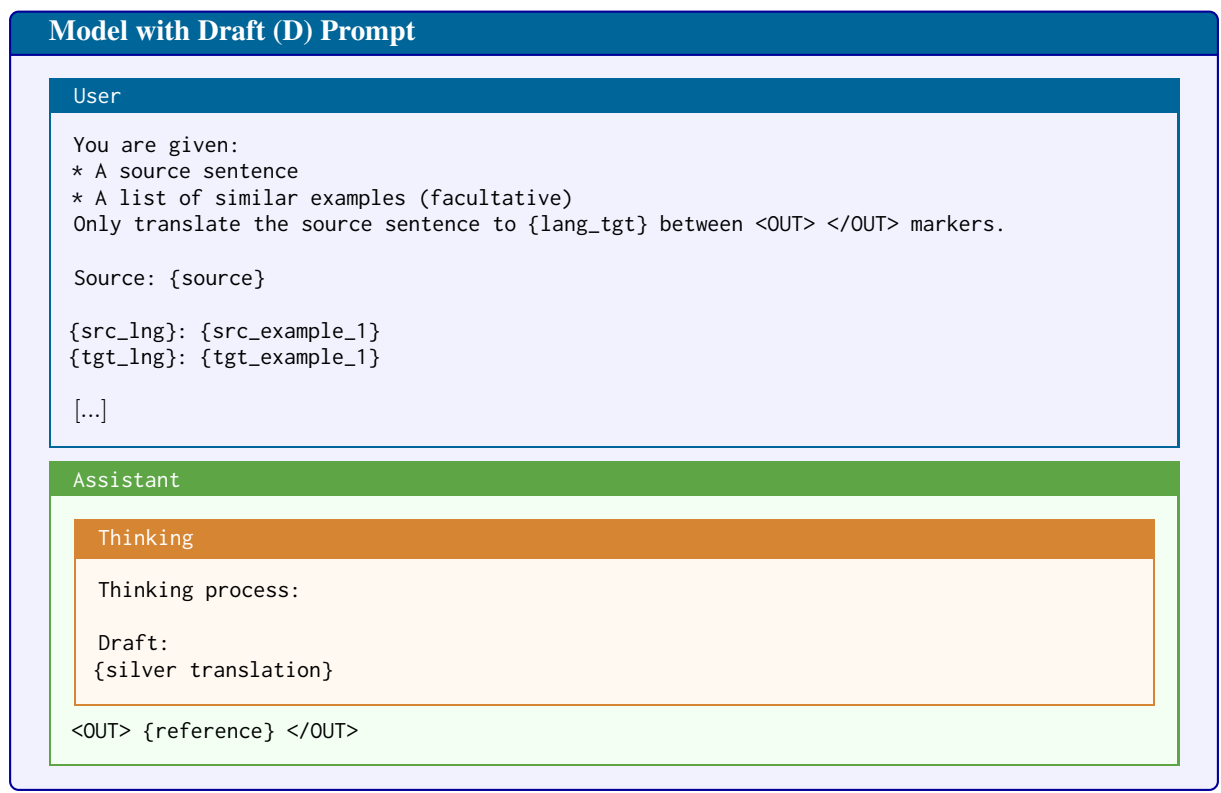

	\centering
	\begin{tcolorbox}[
			title=Model with Draft (D) Prompt,
			coltitle=white,
			colbacktitle=blue!60!green,
			fontupper=\ttfamily\small
		]

		\begin{tcolorbox}[
				colback=blue!5,
				colframe=blue!60!green,
				boxrule=0.8pt,
				title=User,
				sharp corners,
				left=2mm,right=2mm,top=2mm,bottom=2mm
			]
			You are given:\\
			* A source sentence\\
			* A list of similar examples (facultative)\\
			Only translate the source sentence to \{lang\_tgt\} between <OUT> </OUT> markers.\\

			Source: \{source\} \\

			\{src\_lng\}: \{src\_example\_1\} \\
			\{tgt\_lng\}: \{tgt\_example\_1\} \\

			$\left[...\right]$

		\end{tcolorbox}

		\begin{tcolorbox}[
				colback=green!5,
				colframe=green!60!black!70!white!90!red,
				boxrule=0.8pt,
				sharp corners,
				title=Assistant,
				left=2mm,right=2mm,top=2mm,bottom=2mm
			]
			\begin{tcolorbox}[
					colback=orange!5,
					colframe=orange!80!black!80!white,
					boxrule=0.8pt,
					sharp corners,
					title=Thinking,
					left=2mm,right=2mm,top=2mm,bottom=2mm
				]
				Thinking process: \\
				\\
				Draft:\\
				\{silver translation\}
			\end{tcolorbox}
			<OUT>
			\{reference\}
			</OUT>

		\end{tcolorbox}

	\end{tcolorbox}

	\caption{\label{fig:qwen-ft-draft} Prompt used to train the draft model \Draft.}
\end{figure*}

\begin{figure*}
	\centering
	\begin{tcolorbox}[
			title=Model with fragments (F) Prompt,
			coltitle=white,
			colbacktitle=blue!60!green,
			fontupper=\ttfamily\small
		]

		\begin{tcolorbox}[
				colback=blue!5,
				colframe=blue!60!green,
				boxrule=0.8pt,
				title=User,
				sharp corners,
				left=2mm,right=2mm,top=2mm,bottom=2mm
			]
			You are given:\\
			* A source sentence\\
			* A list of similar examples (facultative)\\
			Only translate the source sentence to \{lang\_tgt\} between <OUT> </OUT> markers.\\

			Source: \{source\} \\

			\{src\_lng\}: \{src\_example\_1\} \\
			\{tgt\_lng\}: \{tgt\_example\_1\} \\

			$\left[...\right]$

		\end{tcolorbox}

		\begin{tcolorbox}[
				colback=green!5,
				colframe=green!60!black!70!white!90!red,
				boxrule=0.8pt,
				sharp corners,
				title=Assistant,
				left=2mm,right=2mm,top=2mm,bottom=2mm
			]
			\begin{tcolorbox}[
					colback=orange!5,
					colframe=orange!80!black!80!white,
					boxrule=0.8pt,
					sharp corners,
					title=Thinking,
					left=2mm,right=2mm,top=2mm,bottom=2mm
				]
				Thinking process: \\
				\\
				Fragments:\\
				\{silver fragments\}
			\end{tcolorbox}

			<OUT>
			\{reference\}
			</OUT>

		\end{tcolorbox}

	\end{tcolorbox}

	\caption{\label{fig:qwen-ft-fragments} Prompt used to train the fragments model \Frag.}
\end{figure*}

\begin{figure*}
	\centering
	\begin{tcolorbox}[
			title=Model with Draft and Fragments (F+D) Prompt,
			coltitle=white,
			colbacktitle=blue!60!green,
			fontupper=\ttfamily\small
		]

		\begin{tcolorbox}[
				colback=blue!5,
				colframe=blue!60!green,
				boxrule=0.8pt,
				title=User,
				sharp corners,
				left=2mm,right=2mm,top=2mm,bottom=2mm
			]
			You are given:\\
			* A source sentence\\
			* A list of similar examples (facultative)\\
			Only translate the source sentence to \{lang\_tgt\} between <OUT> </OUT> markers.\\

			Source: \{source\} \\

			\{src\_lng\}: \{src\_example\_1\} \\
			\{tgt\_lng\}: \{tgt\_example\_1\} \\

			$\left[...\right]$

		\end{tcolorbox}

		\begin{tcolorbox}[
				colback=green!5,
				colframe=green!60!black!70!white!90!red,
				boxrule=0.8pt,
				sharp corners,
				title=Assistant,
				left=2mm,right=2mm,top=2mm,bottom=2mm
			]
			\begin{tcolorbox}[
					colback=orange!5,
					colframe=orange!80!black!80!white,
					boxrule=0.8pt,
					sharp corners,
					title=Thinking,
					left=2mm,right=2mm,top=2mm,bottom=2mm
				]
				Thinking process: \\
				\\
				Fragments:\\
				\{silver fragments\}\\
				\\
				Draft:\\
				\{silver translation\}
			\end{tcolorbox}

			<OUT>
			\{reference\}
			</OUT>

		\end{tcolorbox}

	\end{tcolorbox}

	\caption{\label{fig:qwen-ft-draft-fragments} Prompt used to train the draft+fragments model \FragDraft.}
\end{figure*}

\begin{table*}
	\small
	\resizebox{\linewidth}{!}{
		\begin{tabular}{llrrrrrrrrrrrrrrrr}
			\toprule
			          &   & ECB                                       & JRC  & \makebox[\widthof{xxx}]{Wiki} & JRC   & KDE  & TED  & ECB  & \makebox[\widthof{xxx}]{EMEA} & Epp  & JRC  & \makebox[\widthof{xxx}]{Wiki} & KDE  & \makebox[\widthof{xxx}]{EMEA} & JRC  & Kor  & Sub  \\
			think     & k & pl                                        & pl   & pl                            & es    & uk   & uk   & fr   & fr                            & fr   & fr   & fr                            & de   & de                            & de   & de   & de   \\
			\cmidrule(lr){3-18}
			          &   & \multicolumn{16}{c}{\BLEU$\uparrow$}                                                                                                                                                                                                                    \\
			\ding{55} & 0 & 21.2                                      & 27.2 & 28.2                          & 35.3  & 22.1 & 18.7 & 40.0 & 39.5                          & 32.8 & 38.9 & 33.4                          & 30.0 & 33.0                          & 26.1 & 9.8  & 20.9 \\
			\ding{55} & 1 & 32.2                                      & 40.9 & 25.6                          & 43.0  & 25.1 & 17.7 & 48.4 & 48.4                          & 27.8 & 49.1 & 34.6                          & 29.0 & 37.8                          & 36.9 & 13.4 & 18.0 \\
			\ding{55} & 2 & 35.0                                      & 43.4 & 29.0                          & 46.9  & 26.5 & 18.6 & 52.7 & 51.3                          & 31.5 & 52.2 & 36.8                          & 32.0 & 40.4                          & 40.6 & 14.5 & 19.0 \\
			\ding{55} & 3 & 35.5                                      & 43.2 & 28.7                          & 46.8  & 27.7 & 18.7 & 53.4 & 51.1                          & 31.6 & 52.7 & 37.9                          & 32.5 & 40.3                          & 41.0 & 15.1 & 19.2 \\ \cmidrule(lr){3-18}
			          &   & \multicolumn{16}{c}{\COMET$\uparrow$}                                                                                                                                                                                                                   \\
			\ding{55} & 0 & 84.3                                      & 80.1 & 84.2                          & 80.7  & 85.8 & 80.8 & 86.2 & 86.2                          & 86.2 & 86.3 & 83.8                          & 79.9 & 81.9                          & 83.6 & 71.0 & 76.2 \\
			\ding{55} & 1 & 84.5                                      & 83.6 & 80.8                          & 79.4  & 83.3 & 77.3 & 83.5 & 85.8                          & 81.7 & 84.9 & 81.8                          & 74.5 & 80.0                          & 81.3 & 66.1 & 70.0 \\
			\ding{55} & 2 & 85.9                                      & 85.1 & 83.5                          & 81.1  & 85.2 & 79.4 & 85.7 & 86.8                          & 84.5 & 86.1 & 83.9                          & 76.7 & 81.3                          & 83.2 & 69.3 & 73.1 \\
			\ding{55} & 3 & 86.0                                      & 84.7 & 83.3                          & 81.3  & 86.0 & 79.3 & 86.5 & 86.8                          & 84.4 & 86.0 & 84.1                          & 77.3 & 81.6                          & 83.4 & 69.2 & 73.5 \\ \cmidrule(lr){3-18}
			          &   & \multicolumn{16}{c}{\MetricX$\downarrow$}                                                                                                                                                                                                               \\
			\ding{55} & 0 & 7.22                                      & 9.35 & 7.33                          & 10.75 & 3.46 & 6.40 & 8.43 & 6.03                          & 7.43 & 9.29 & 9.20                          & 3.91 & 7.16                          & 9.15 & 9.09 & 4.76 \\
			\ding{55} & 1 & 2.05                                      & 2.60 & 3.35                          & 1.92  & 1.94 & 4.36 & 1.83 & 1.40                          & 1.73 & 1.38 & 2.53                          & 1.49 & 1.54                          & 1.13 & 3.27 & 1.91 \\
			\ding{55} & 2 & 1.96                                      & 2.55 & 3.28                          & 1.89  & 1.97 & 4.42 & 1.78 & 1.34                          & 1.70 & 1.35 & 2.52                          & 1.47 & 1.53                          & 1.08 & 3.26 & 1.88 \\
			\ding{55} & 3 & 1.98                                      & 2.46 & 3.20                          & 1.85  & 1.96 & 4.40 & 1.76 & 1.33                          & 1.75 & 1.32 & 2.52                          & 1.47 & 1.53                          & 1.09 & 3.20 & 1.87 \\
			\midrule
			\bottomrule
		\end{tabular}
	}
	\caption{\label{tab:model-I} \BLEU, \COMET, and MetricX scores on the multi-domain test sets for the pre-trained Qwen3-8B Instruct model \Instruct.}
\end{table*}

\begin{table*}
	\small
	\resizebox{\linewidth}{!}{
		\begin{tabular}{llrrrrrrrrrrrrrrrr}
			\toprule
			          &   & ECB                                  & JRC  & \makebox[\widthof{xxx}]{Wiki} & JRC  & KDE  & TED  & ECB  & \makebox[\widthof{xxx}]{EMEA} & Epp  & JRC  & \makebox[\widthof{xxx}]{Wiki} & KDE  & \makebox[\widthof{xxx}]{EMEA} & JRC  & Kor  & Sub  \\
			think     & k & pl                                   & pl   & pl                            & es   & uk   & uk   & fr   & fr                            & fr   & fr   & fr                            & de   & de                            & de   & de   & de   \\
			\cmidrule(lr){3-18}
			          &   & \multicolumn{16}{c}{\BLEU$\uparrow$}                                                                                                                                                                                                              \\
			\ding{55} & 0 & 40.8                                 & 44.5 & 36.5                          & 50.4 & 37.6 & 21.4 & 51.3 & 49.0                          & 36.1 & 52.2 & 40.3                          & 34.5 & 39.4                          & 37.0 & 19.4 & 23.1 \\
			\ding{55} & 1 & 48.3                                 & 54.7 & 38.0                          & 60.7 & 42.9 & 21.7 & 63.6 & 64.0                          & 36.6 & 65.8 & 43.4                          & 41.1 & 47.2                          & 53.6 & 22.2 & 20.9 \\
			\ding{55} & 2 & 48.8                                 & 55.2 & 38.2                          & 61.3 & 43.3 & 21.6 & 64.8 & 64.9                          & 37.0 & 67.1 & 43.7                          & 41.7 & 47.8                          & 54.3 & 23.3 & 23.5 \\
			\ding{55} & 3 & 49.0                                 & 55.5 & 38.6                          & 61.9 & 43.4 & 21.7 & 65.2 & 65.2                          & 36.9 & 67.5 & 43.9                          & 42.2 & 47.8                          & 54.8 & 24.0 & 23.6 \\ \cmidrule(lr){3-18} && \multicolumn{16}{c}{\COMET$\uparrow$} \\
			\ding{55} & 0 & 89.6                                 & 88.0 & 88.2                          & 84.5 & 89.8 & 83.9 & 88.2 & 88.3                          & 86.8 & 88.5 & 85.8                          & 82.7 & 84.0                          & 85.7 & 72.8 & 78.1 \\
			\ding{55} & 1 & 90.6                                 & 89.8 & 88.3                          & 85.7 & 90.5 & 83.9 & 89.9 & 90.0                          & 86.9 & 90.1 & 86.5                          & 84.0 & 85.1                          & 87.7 & 73.2 & 78.3 \\
			\ding{55} & 2 & 90.4                                 & 90.0 & 88.5                          & 85.8 & 90.6 & 84.0 & 90.1 & 90.2                          & 87.0 & 90.3 & 86.6                          & 84.2 & 85.2                          & 88.0 & 73.5 & 78.3 \\
			\ding{55} & 3 & 90.5                                 & 90.3 & 88.6                          & 85.9 & 90.7 & 84.0 & 90.2 & 90.2                          & 87.0 & 90.4 & 86.7                          & 84.3 & 85.3                          & 88.1 & 73.6 & 78.5 \\ \cmidrule(lr){3-18} && \multicolumn{16}{c}{\MetricX$\downarrow$} \\
			\ding{55} & 0 & 2.13                                 & 3.22 & 3.24                          & 2.04 & 2.02 & 4.49 & 2.01 & 1.55                          & 1.76 & 1.57 & 2.57                          & 1.59 & 1.62                          & 1.31 & 3.27 & 1.94 \\
			\ding{55} & 1 & 1.85                                 & 2.49 & 3.21                          & 1.85 & 1.94 & 4.45 & 1.78 & 1.36                          & 1.75 & 1.35 & 2.47                          & 1.48 & 1.53                          & 1.11 & 3.28 & 1.89 \\
			\ding{55} & 2 & 1.86                                 & 2.43 & 3.18                          & 1.80 & 1.94 & 4.45 & 1.73 & 1.35                          & 1.72 & 1.32 & 2.48                          & 1.46 & 1.51                          & 1.09 & 3.23 & 1.89 \\
			\ding{55} & 3 & 1.84                                 & 2.31 & 3.10                          & 1.79 & 1.93 & 4.44 & 1.72 & 1.34                          & 1.72 & 1.31 & 2.45                          & 1.46 & 1.49                          & 1.07 & 3.23 & 1.89 \\
			\midrule
			\bottomrule
		\end{tabular}
	}
	\caption{\label{tab:model-B} \BLEU, \COMET, and MetricX scores on the multi-domain test sets for the fine-tuned Qwen3-8B baseline model \Baseline.}
\end{table*}

\begin{table*}
	\small
	\resizebox{\linewidth}{!}{
		\begin{tabular}{llrrrrrrrrrrrrrrrr}
			\toprule
			                  &   & ECB                                  & JRC  & \makebox[\widthof{xxx}]{Wiki} & JRC  & KDE  & TED  & ECB  & \makebox[\widthof{xxx}]{EMEA} & Epp  & JRC  & \makebox[\widthof{xxx}]{Wiki} & KDE  & \makebox[\widthof{xxx}]{EMEA} & JRC  & Kor  & Sub  \\
			think             & k & pl                                   & pl   & pl                            & es   & uk   & uk   & fr   & fr                            & fr   & fr   & fr                            & de   & de                            & de   & de   & de   \\
			\cmidrule(lr){3-18}
			                  &   & \multicolumn{16}{c}{\BLEU$\uparrow$}                                                                                                                                                                                                              \\
			\ding{55}         & 0 & 36.3                                 & 40.5 & 34.6                          & 48.0 & 34.8 & 21.2 & 49.6 & 47.5                          & 35.7 & 50.5 & 39.4                          & 33.9 & 39.2                          & 36.4 & 17.6 & 21.1 \\
			\ding{55}         & 1 & 46.3                                 & 53.2 & 36.9                          & 59.4 & 40.9 & 21.8 & 63.3 & 63.6                          & 36.2 & 64.8 & 43.5                          & 41.1 & 47.6                          & 52.7 & 20.6 & 23.7 \\
			\ding{55}         & 2 & 47.0                                 & 54.0 & 37.5                          & 60.3 & 41.3 & 21.6 & 64.7 & 64.7                          & 36.3 & 66.7 & 44.0                          & 41.6 & 48.1                          & 54.2 & 21.6 & 24.0 \\
			\ding{55}         & 3 & 47.5                                 & 54.5 & 37.8                          & 60.7 & 41.5 & 21.8 & 65.1 & 64.9                          & 36.6 & 67.1 & 44.2                          & 41.7 & 48.4                          & 54.6 & 22.7 & 24.3 \\
			\green{\ding{51}} & 0 & 37.0                                 & 40.7 & 36.0                          & 48.3 & 34.5 & 21.4 & 49.5 & 46.3                          & 35.7 & 49.9 & 40.4                          & 34.1 & 38.9                          & 36.9 & 17.0 & 22.9 \\
			\green{\ding{51}} & 1 & 46.8                                 & 53.1 & 36.8                          & 59.3 & 41.1 & 21.6 & 62.0 & 62.4                          & 34.5 & 63.9 & 41.1                          & 41.1 & 47.2                          & 52.9 & 20.8 & 21.9 \\
			\green{\ding{51}} & 2 & 47.4                                 & 53.7 & 37.5                          & 60.4 & 40.9 & 21.8 & 63.4 & 64.1                          & 34.4 & 65.2 & 43.0                          & 41.8 & 48.1                          & 54.0 & 21.3 & 23.7 \\
			\green{\ding{51}} & 3 & 47.9                                 & 54.4 & 38.1                          & 61.2 & 41.8 & 21.8 & 65.0 & 65.1                          & 36.6 & 67.8 & 44.0                          & 42.2 & 48.4                          & 54.5 & 23.0 & 23.7 \\ \cmidrule(lr){3-18} && \multicolumn{16}{c}{\COMET$\uparrow$} \\
			\ding{55}         & 0 & 88.5                                 & 87.0 & 87.7                          & 84.0 & 88.9 & 83.6 & 88.2 & 88.1                          & 86.8 & 88.2 & 86.1                          & 82.5 & 83.8                          & 85.5 & 72.9 & 77.9 \\
			\ding{55}         & 1 & 89.9                                 & 89.3 & 88.2                          & 85.4 & 90.1 & 83.7 & 89.9 & 90.1                          & 86.9 & 90.0 & 86.6                          & 83.6 & 85.0                          & 87.6 & 73.3 & 77.9 \\
			\ding{55}         & 2 & 90.0                                 & 89.5 & 88.3                          & 85.5 & 90.3 & 83.7 & 90.1 & 90.2                          & 86.9 & 90.3 & 86.7                          & 83.9 & 85.1                          & 87.9 & 73.4 & 78.1 \\
			\ding{55}         & 3 & 90.2                                 & 89.8 & 88.4                          & 85.5 & 90.4 & 84.0 & 90.1 & 90.2                          & 87.0 & 90.4 & 86.7                          & 84.0 & 85.2                          & 88.0 & 73.7 & 78.2 \\
			\green{\ding{51}} & 0 & 88.8                                 & 86.9 & 88.2                          & 84.1 & 88.9 & 82.8 & 88.0 & 88.1                          & 87.0 & 88.2 & 85.9                          & 82.4 & 83.7                          & 85.6 & 72.8 & 77.1 \\
			\green{\ding{51}} & 1 & 90.0                                 & 89.1 & 87.7                          & 85.1 & 90.1 & 82.7 & 88.7 & 89.0                          & 85.0 & 88.4 & 83.0                          & 83.2 & 84.6                          & 87.1 & 72.1 & 74.7 \\
			\green{\ding{51}} & 2 & 90.2                                 & 89.1 & 88.1                          & 85.5 & 90.2 & 82.9 & 89.5 & 90.0                          & 85.0 & 89.1 & 84.9                          & 83.9 & 85.0                          & 87.4 & 72.3 & 77.2 \\
                  \green{\ding{51}} & 3 & 90.2                                 & 89.1 & 88.1                          & 85.5 & 90.2 & 82.9 & 89.5 & 90.0                          & 85.0 & 89.1 & 84.9                          & 83.9 & 85.0                          & 87.4 & 72.3 & 77.2 \\
                  \cmidrule(lr){3-18} && \multicolumn{16}{c}{\MetricX$\downarrow$} \\
			\ding{55}         & 0 & 2.52                                 & 3.56 & 3.51                          & 2.16 & 2.23 & 4.66 & 1.98 & 1.60                          & 1.78 & 1.63 & 2.51                          & 1.61 & 1.66                          & 1.29 & 3.09 & 1.91 \\
			\ding{55}         & 1 & 2.11                                 & 2.73 & 3.29                          & 1.90 & 2.04 & 4.63 & 1.77 & 1.35                          & 1.75 & 1.39 & 2.46                          & 1.54 & 1.54                          & 1.09 & 3.12 & 1.92 \\
			\ding{55}         & 2 & 2.07                                 & 2.59 & 3.22                          & 1.86 & 2.01 & 4.59 & 1.74 & 1.33                          & 1.72 & 1.36 & 2.50                          & 1.49 & 1.53                          & 1.07 & 3.12 & 1.91 \\
			\ding{55}         & 3 & 1.99                                 & 2.50 & 3.18                          & 1.86 & 2.00 & 4.54 & 1.75 & 1.32                          & 1.71 & 1.34 & 2.50                          & 1.48 & 1.53                          & 1.07 & 3.08 & 1.89 \\
			\green{\ding{51}} & 0 & 2.46                                 & 3.53 & 3.16                          & 2.14 & 2.23 & 4.87 & 2.05 & 1.58                          & 1.72 & 1.64 & 2.41                          & 1.63 & 1.66                          & 1.27 & 3.05 & 2.04 \\
			\green{\ding{51}} & 1 & 2.10                                 & 2.75 & 3.41                          & 2.07 & 2.01 & 5.01 & 2.31 & 1.79                          & 2.72 & 2.07 & 4.70                          & 1.66 & 1.73                          & 1.41 & 3.67 & 3.17 \\
			\green{\ding{51}} & 2 & 2.04                                 & 2.66 & 3.25                          & 1.90 & 2.02 & 4.94 & 2.07 & 1.44                          & 2.78 & 1.82 & 3.55                          & 1.52 & 1.59                          & 1.29 & 3.74 & 2.28 \\
			\green{\ding{51}} & 3 & 1.98                                 & 2.56 & 3.04                          & 1.83 & 1.99 & 4.73 & 1.74 & 1.32                          & 1.68 & 1.34 & 2.40                          & 1.50 & 1.54                          & 1.05 & 3.13 & 1.90 \\		\midrule
			\bottomrule
		\end{tabular}
	}
	\caption{\label{tab:model-D} \BLEU, \COMET, and MetricX scores on the multi-domain test sets for the fine-tuned Qwen3-8B drafting model \Draft.}
\end{table*}

\begin{table*}
	\small
	\resizebox{\linewidth}{!}{
		\begin{tabular}{llrrrrrrrrrrrrrrrr}
			\toprule
			                  &   & ECB                                  & JRC  & \makebox[\widthof{xxx}]{Wiki} & JRC  & KDE  & TED  & ECB  & \makebox[\widthof{xxx}]{EMEA} & Epp  & JRC  & \makebox[\widthof{xxx}]{Wiki} & KDE  & \makebox[\widthof{xxx}]{EMEA} & JRC  & Kor  & Sub  \\
			think             & k & pl                                   & pl   & pl                            & es   & uk   & uk   & fr   & fr                            & fr   & fr   & fr                            & de   & de                            & de   & de   & de   \\
			\cmidrule(lr){3-18}
			                  &   & \multicolumn{16}{c}{\BLEU$\uparrow$}                                                                                                                                                                                                              \\
			\ding{55}         & 0 & 38.0                                 & 42.5 & 35.4                          & 49.1 & 36.5 & 21.5 & 50.8 & 47.9                          & 35.8 & 51.4 & 36.5                          & 33.8 & 39.0                          & 37.3 & 18.9 & 23.2 \\
			\ding{55}         & 1 & 46.5                                 & 53.9 & 37.3                          & 59.5 & 41.7 & 21.6 & 63.1 & 63.5                          & 36.2 & 65.0 & 43.4                          & 40.8 & 47.0                          & 52.7 & 21.8 & 23.3 \\
			\ding{55}         & 2 & 47.4                                 & 54.7 & 37.7                          & 60.7 & 42.3 & 21.6 & 64.3 & 64.2                          & 36.2 & 66.6 & 43.9                          & 41.3 & 47.9                          & 54.1 & 22.5 & 23.4 \\
			\ding{55}         & 3 & 48.1                                 & 55.1 & 38.0                          & 61.2 & 42.3 & 21.5 & 64.9 & 64.7                          & 36.2 & 66.9 & 44.2                          & 41.7 & 48.0                          & 54.8 & 23.4 & 23.5 \\
			\green{\ding{51}} & 0 & 39.2                                 & 43.5 & 36.5                          & 49.9 & 37.2 & 22.5 & 51.3 & 49.9                          & 36.6 & 52.1 & 41.7                          & 34.7 & 40.1                          & 38.4 & 19.1 & 23.8 \\
			\green{\ding{51}} & 1 & 48.0                                 & 54.2 & 38.3                          & 60.1 & 43.1 & 22.7 & 63.9 & 64.4                          & 37.3 & 64.1 & 43.8                          & 41.9 & 47.7                          & 53.8 & 22.8 & 24.3 \\
			\green{\ding{51}} & 2 & 48.6                                 & 55.5 & 38.7                          & 61.2 & 43.6 & 22.8 & 64.8 & 65.2                          & 37.4 & 67.6 & 43.9                          & 42.4 & 48.6                          & 54.7 & 23.7 & 24.3 \\
			\green{\ding{51}} & 3 & 49.0                                 & 55.6 & 38.7                          & 62.0 & 43.6 & 22.8 & 65.2 & 65.5                          & 37.4 & 67.9 & 44.1                          & 42.5 & 48.8                          & 55.2 & 24.1 & 24.2 \\ \cmidrule(lr){3-18} && \multicolumn{16}{c}{\COMET$\uparrow$} \\
			\ding{55}         & 0 & 88.9                                 & 87.5 & 87.8                          & 84.2 & 89.5 & 84.0 & 88.2 & 88.2                          & 86.8 & 88.3 & 84.6                          & 82.5 & 83.8                          & 85.7 & 73.1 & 78.0 \\
			\ding{55}         & 1 & 90.2                                 & 89.6 & 88.1                          & 85.5 & 90.3 & 84.0 & 89.8 & 90.0                          & 86.9 & 90.1 & 86.5                          & 83.7 & 84.8                          & 87.6 & 73.5 & 78.2 \\
			\ding{55}         & 2 & 90.3                                 & 89.8 & 88.4                          & 85.7 & 90.4 & 84.3 & 89.9 & 90.1                          & 87.0 & 90.3 & 86.6                          & 84.1 & 85.0                          & 87.8 & 73.5 & 78.2 \\
			\ding{55}         & 3 & 90.4                                 & 89.9 & 88.4                          & 85.7 & 90.5 & 84.1 & 90.0 & 90.0                          & 87.0 & 90.4 & 86.6                          & 84.2 & 85.1                          & 88.1 & 73.8 & 78.3 \\
			\green{\ding{51}} & 0 & 89.5                                 & 88.1 & 88.7                          & 84.5 & 89.9 & 84.7 & 88.5 & 88.8                          & 87.2 & 88.5 & 86.3                          & 82.7 & 84.2                          & 86.2 & 73.3 & 78.6 \\
			\green{\ding{51}} & 1 & 90.7                                 & 89.8 & 88.9                          & 85.6 & 90.7 & 84.7 & 89.8 & 90.0                          & 86.9 & 90.1 & 86.5                          & 83.7 & 84.8                          & 87.6 & 73.5 & 78.2 \\
			\green{\ding{51}} & 2 & 90.7                                 & 90.1 & 89.0                          & 85.9 & 90.8 & 85.0 & 90.1 & 90.3                          & 87.3 & 90.4 & 86.8                          & 84.2 & 85.3                          & 88.1 & 74.2 & 78.6 \\
                  \green{\ding{51}} & 3 & 90.8                                 & 90.2 & 89.0                          & 86.0 & 90.9 & 84.9 & 90.3 & 90.3                          & 87.3 & 90.5 & 86.8                          & 84.3 & 85.2                          & 88.3 & 74.1 & 78.7 \\
                  \cmidrule(lr){3-18} && \multicolumn{16}{c}{\MetricX$\downarrow$} \\
			\ding{55}         & 0 & 2.40                                 & 3.33 & 3.36                          & 2.13 & 2.14 & 4.41 & 2.02 & 1.55                          & 1.78 & 1.60 & 2.94                          & 1.62 & 1.63                          & 1.30 & 3.13 & 1.91 \\
			\ding{55}         & 1 & 2.06                                 & 2.55 & 3.31                          & 1.88 & 2.00 & 4.42 & 1.79 & 1.37                          & 1.74 & 1.33 & 2.46                          & 1.51 & 1.55                          & 1.10 & 3.12 & 1.89 \\
			\ding{55}         & 2 & 2.01                                 & 2.49 & 3.26                          & 1.86 & 2.00 & 4.34 & 1.77 & 1.34                          & 1.72 & 1.31 & 2.49                          & 1.48 & 1.52                          & 1.08 & 3.12 & 1.90 \\
			\ding{55}         & 3 & 1.92                                 & 2.40 & 3.28                          & 1.82 & 1.95 & 4.43 & 1.76 & 1.37                          & 1.72 & 1.32 & 2.47                          & 1.47 & 1.52                          & 1.05 & 3.08 & 1.88 \\
			\green{\ding{51}} & 0 & 2.11                                 & 3.11 & 2.89                          & 2.03 & 2.01 & 4.16 & 1.93 & 1.48                          & 1.66 & 1.54 & 2.37                          & 1.57 & 1.59                          & 1.21 & 3.05 & 1.80 \\
			\green{\ding{51}} & 1 & 1.88                                 & 2.46 & 2.85                          & 1.85 & 1.89 & 4.17 & 1.74 & 1.34                          & 1.63 & 1.35 & 2.38                          & 1.47 & 1.51                          & 1.08 & 3.03 & 1.79 \\
			\green{\ding{51}} & 2 & 1.80                                 & 2.33 & 2.80                          & 1.77 & 1.88 & 4.11 & 1.73 & 1.30                          & 1.62 & 1.31 & 2.35                          & 1.44 & 1.48                          & 1.03 & 2.96 & 1.81 \\
			\green{\ding{51}} & 3 & 1.76                                 & 2.32 & 2.85                          & 1.76 & 1.85 & 4.16 & 1.71 & 1.30                          & 1.62 & 1.30 & 2.36                          & 1.44 & 1.49                          & 1.02 & 3.03 & 1.80 \\		\midrule
			\bottomrule
		\end{tabular}
	}
	\caption{\label{tab:model-F} \BLEU, \COMET, and MetricX scores on the multi-domain test sets for the fine-tuned Qwen3-8B fragment-based model \Frag.}
\end{table*}

\begin{table*}
	\small
	\resizebox{\linewidth}{!}{
		\begin{tabular}{llrrrrrrrrrrrrrrrr}
			\toprule
			                  &   & ECB                                  & JRC  & \makebox[\widthof{xxx}]{Wiki} & JRC  & KDE  & TED  & ECB  & \makebox[\widthof{xxx}]{EMEA} & Epp  & JRC  & \makebox[\widthof{xxx}]{Wiki} & KDE  & \makebox[\widthof{xxx}]{EMEA} & JRC  & Kor  & Sub  \\
			think             & k & pl                                   & pl   & pl                            & es   & uk   & uk   & fr   & fr                            & fr   & fr   & fr                            & de   & de                            & de   & de   & de   \\
			\cmidrule(lr){3-18}
			                  &   & \multicolumn{16}{c}{\BLEU$\uparrow$}                                                                                                                                                                                                              \\
			\ding{55}         & 0 & 37.9                                 & 41.8 & 35.1                          & 49.0 & 35.9 & 21.1 & 50.2 & 49.0                          & 35.5 & 51.2 & 41.2                          & 33.9 & 39.2                          & 37.0 & 18.6 & 23.2 \\
			\ding{55}         & 1 & 46.4                                 & 53.7 & 37.1                          & 59.6 & 41.3 & 21.3 & 63.1 & 63.0                          & 35.9 & 64.3 & 43.5                          & 41.3 & 46.7                          & 52.4 & 21.7 & 23.4 \\
			\ding{55}         & 2 & 47.4                                 & 54.5 & 37.4                          & 60.6 & 42.0 & 21.3 & 64.2 & 64.3                          & 36.1 & 66.3 & 43.8                          & 41.6 & 47.7                          & 53.9 & 22.7 & 23.7 \\
			\ding{55}         & 3 & 47.8                                 & 55.0 & 37.8                          & 61.1 & 42.0 & 21.4 & 64.8 & 64.5                          & 36.1 & 66.7 & 43.9                          & 41.6 & 48.1                          & 54.3 & 23.3 & 23.6 \\
			\green{\ding{51}} & 0 & 39.1                                 & 42.6 & 36.6                          & 50.1 & 36.9 & 22.0 & 49.6 & 49.3                          & 36.7 & 52.1 & 41.8                          & 34.6 & 40.0                          & 37.9 & 18.6 & 23.9 \\
			\green{\ding{51}} & 1 & 47.9                                 & 54.3 & 37.9                          & 60.3 & 42.8 & 22.2 & 64.1 & 64.2                          & 37.4 & 65.8 & 44.0                          & 41.8 & 47.8                          & 53.4 & 22.5 & 24.1 \\
			\green{\ding{51}} & 2 & 48.3                                 & 55.3 & 38.6                          & 61.2 & 43.5 & 22.4 & 64.9 & 64.7                          & 37.4 & 67.5 & 44.5                          & 42.5 & 48.3                          & 54.5 & 23.3 & 24.2 \\
			\green{\ding{51}} & 3 & 48.7                                 & 55.6 & 38.8                          & 62.1 & 43.5 & 22.4 & 65.4 & 65.3                          & 37.6 & 67.7 & 44.4                          & 42.9 & 48.0                          & 55.0 & 24.3 & 24.3 \\ \cmidrule(lr){3-18} && \multicolumn{16}{c}{\COMET$\uparrow$} \\
			\ding{55}         & 0 & 88.6                                 & 87.4 & 87.7                          & 84.1 & 89.4 & 83.4 & 87.9 & 88.1                          & 86.8 & 88.3 & 85.8                          & 82.4 & 83.7                          & 85.5 & 73.1 & 78.1 \\
			\ding{55}         & 1 & 90.2                                 & 89.6 & 88.1                          & 85.5 & 90.2 & 83.7 & 89.7 & 89.8                          & 86.9 & 89.8 & 86.5                          & 83.7 & 84.9                          & 87.6 & 73.5 & 78.1 \\
			\ding{55}         & 2 & 90.4                                 & 89.8 & 88.2                          & 85.6 & 90.4 & 83.7 & 90.0 & 90.0                          & 86.9 & 90.1 & 86.7                          & 83.9 & 85.1                          & 87.9 & 73.8 & 78.2 \\
			\ding{55}         & 3 & 90.4                                 & 90.0 & 88.4                          & 85.7 & 90.5 & 83.9 & 90.0 & 90.1                          & 87.0 & 90.3 & 86.6                          & 84.1 & 85.1                          & 88.0 & 73.7 & 78.2 \\
			\green{\ding{51}} & 0 & 89.3                                 & 87.9 & 88.6                          & 84.4 & 89.8 & 84.2 & 88.4 & 88.6                          & 87.0 & 88.6 & 86.3                          & 82.7 & 84.2                          & 86.2 & 73.4 & 78.3 \\
			\green{\ding{51}} & 1 & 90.6                                 & 89.8 & 88.8                          & 85.7 & 90.6 & 84.4 & 90.0 & 90.1                          & 87.2 & 90.2 & 86.6                          & 84.0 & 85.2                          & 87.9 & 73.7 & 78.6 \\
			\green{\ding{51}} & 2 & 90.7                                 & 90.0 & 89.0                          & 85.8 & 90.8 & 84.5 & 90.1 & 90.2                          & 87.1 & 90.4 & 86.7                          & 84.1 & 85.3                          & 88.1 & 73.6 & 78.6 \\
                  \green{\ding{51}} & 3 & 90.7                                 & 90.0 & 88.9                          & 86.0 & 90.8 & 84.5 & 90.2 & 90.3                          & 87.2 & 90.5 & 86.8                          & 84.3 & 85.3                          & 88.2 & 73.9 & 78.7 \\
                  \cmidrule(lr){3-18} && \multicolumn{16}{c}{\MetricX$\downarrow$} \\
			\ding{55}         & 0 & 2.39                                 & 3.39 & 3.53                          & 2.14 & 2.19 & 4.59 & 2.11 & 1.59                          & 1.74 & 1.61 & 2.58                          & 1.61 & 1.66                          & 1.33 & 3.14 & 1.91 \\
			\ding{55}         & 1 & 1.99                                 & 2.58 & 3.39                          & 1.89 & 2.03 & 4.58 & 1.81 & 1.37                          & 1.74 & 1.41 & 2.43                          & 1.51 & 1.56                          & 1.11 & 3.11 & 1.91 \\
			\ding{55}         & 2 & 1.93                                 & 2.51 & 3.24                          & 1.84 & 2.00 & 4.55 & 1.74 & 1.33                          & 1.72 & 1.35 & 2.45                          & 1.49 & 1.53                          & 1.08 & 3.06 & 1.91 \\
			\ding{55}         & 3 & 1.91                                 & 2.43 & 3.20                          & 1.84 & 2.00 & 4.48 & 1.74 & 1.33                          & 1.72 & 1.34 & 2.49                          & 1.48 & 1.53                          & 1.09 & 3.03 & 1.92 \\
			\green{\ding{51}} & 0 & 2.14                                 & 3.21 & 3.01                          & 2.05 & 2.07 & 4.34 & 1.94 & 1.50                          & 1.68 & 1.53 & 2.36                          & 1.55 & 1.59                          & 1.25 & 3.09 & 1.84 \\
			\green{\ding{51}} & 1 & 1.82                                 & 2.50 & 2.90                          & 1.84 & 1.93 & 4.26 & 1.74 & 1.35                          & 1.64 & 1.34 & 2.37                          & 1.48 & 1.53                          & 1.06 & 3.04 & 1.80 \\
			\green{\ding{51}} & 2 & 1.77                                 & 2.41 & 2.81                          & 1.80 & 1.91 & 4.26 & 1.72 & 1.31                          & 1.66 & 1.31 & 2.38                          & 1.46 & 1.50                          & 1.04 & 3.09 & 1.81 \\
			\green{\ding{51}} & 3 & 1.78                                 & 2.40 & 2.90                          & 1.77 & 1.90 & 4.28 & 1.71 & 1.29                          & 1.63 & 1.30 & 2.34                          & 1.45 & 1.50                          & 1.02 & 3.02 & 1.83 \\		\midrule
			\bottomrule
		\end{tabular}
	}
	\caption{\label{tab:model-FD} \BLEU, \COMET, and MetricX scores on the multi-domain test sets for the fine-tuned Qwen3-8B fragment-based drafting model \FragDraft.}
\end{table*}

\section{Tracing Translation \label{appendix:traceability}}

\begin{figure*}
	\centering
	\includegraphics[width=\textwidth]{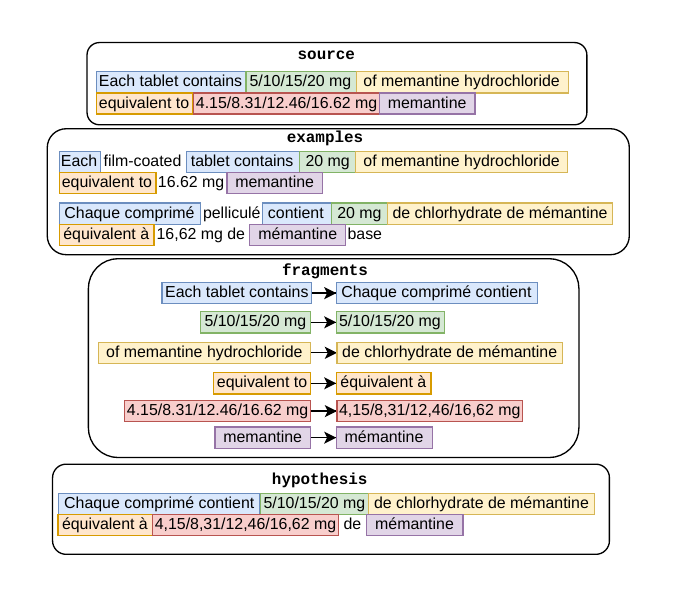}
	\caption{\label{fig:traceability}Illustration of the traceability provided by the FE step.}
\end{figure*}

Figure~\ref{fig:traceability} illustrates the enhanced traceability provided by the fragment-based translation process. We can trace back spans in the generated translations to both the source sentence and the retrieved exemplars.

\end{document}